\pdfoutput=1

\documentclass[11pt]{article}

\usepackage[]{acl}
\usepackage{times}
\usepackage{booktabs}
\usepackage{latexsym}
\usepackage{amssymb}
\usepackage{amsmath}
\usepackage{listings}
\usepackage{pifont}
\usepackage{enumitem}
\usepackage[T1]{fontenc}
\usepackage[utf8]{inputenc}
\usepackage{microtype}
\usepackage{inconsolata}
\usepackage{hyperref}
\usepackage{url}
\usepackage{wrapfig}
\usepackage{graphicx}
\usepackage{lipsum}
\usepackage{booktabs} 
\usepackage{microtype}
\usepackage{inconsolata}
\usepackage{amssymb}
\usepackage{booktabs}
\usepackage{hyperref}
\usepackage{colortbl}
\usepackage{caption}
\usepackage[most]{tcolorbox}
\usepackage{amsmath}
\usepackage{multirow}
\usepackage{color}
\usepackage{pifont}       
\usepackage{bbding}
\usepackage{pifont}
\usepackage{algorithm}
\usepackage{algpseudocode}
\definecolor{deepred}{HTML}{FC8D59}
\definecolor{medred}{HTML}{FDC8B0}
\definecolor{lightred}{HTML}{F5E1D9} 
\definecolor{verylightred}{HTML}{F5E1D9}
\definecolor{extremelylightred}{HTML}{FCF4EF}
\definecolor{lightblue}{HTML}{E4F0F7}
\definecolor{medblue}{HTML}{CBE1EE}
\usepackage{mdframed}
\definecolor{green}{HTML}{65A542}
\usepackage{color}
\definecolor{shadecolor}{rgb}{0.92,0.92,0.92}

\usepackage{hyperref}
\usepackage{xcolor}

\title{Disentangling Language and Culture for Evaluating \\ Multilingual Large Language Models}

\author{
Jiahao Ying~\textsuperscript{1, 5}\thanks{This work was performed when Jiahao Ying, Wei Tang and Yiran Zhao were interns at Alibaba DAMO Academy.}, 
Wei Tang~\textsuperscript{2, 5}, 
Yiran Zhao~\textsuperscript{3, 5},  
Yixin Cao~\textsuperscript{4}, 
Yu Rong~\textsuperscript{5}, 
\textbf{Wenxuan Zhang}~\textsuperscript{6}\thanks{Corresponding author.} 
\vspace{2mm} \\ 
\textsuperscript{1} Singapore Management University \;  \\
\textsuperscript{2} University of Science and Technology of China \\
\textsuperscript{3} National University of Singapore \; \\
\textsuperscript{4} Institute of Trustworthy Embodied AI, Fudan University  \\
\textsuperscript{5} DAMO Academy, Alibaba Group   \; \\
\textsuperscript{6} Singapore University of Technology and Design   \\
}

\begin{document}
\maketitle

\begin{abstract}
    This paper introduces a Dual Evaluation Framework to comprehensively assess the multilingual capabilities of LLMs. By decomposing the evaluation along the dimensions of linguistic medium and cultural context, this framework enables a nuanced analysis of LLMs’ ability to process questions within both native and cross-cultural contexts cross-lingually. Extensive evaluations are conducted on a wide range of models, revealing a notable “Cultural-Linguistic Synergy” phenomenon, where models exhibit better performance when questions are culturally aligned with the language. This phenomenon is further explored through interpretability probing, which shows that a higher proportion of specific neurons are activated in a language’s cultural context. This activation proportion could serve as a potential indicator for evaluating multilingual performance during model training. Our findings challenge the prevailing notion that LLMs, primarily trained on English data, perform uniformly across languages and highlight the necessity of culturally and linguistically model evaluations. Our code can be found at \url{https://yingjiahao14.github.io/Dual-Evaluation/}.
\end{abstract}

\section{Introduction}

With the rapid development of large language models (LLMs), increasing efforts have been made to make these models beneficial for people worldwide. To achieve this, non-English corpora are also incorporated into the training data, enabling LLMs to understand and generate text in various languages (i.e., multilingual capabilities)~\cite{xue-etal-2021-mt5,  grattafiori2024llama3herdmodels, openai2024gpt4technicalreport, nguyen-etal-2024-seallms,seallms3}. 

\begin{figure}[t]
\centering
\includegraphics[scale=0.23]{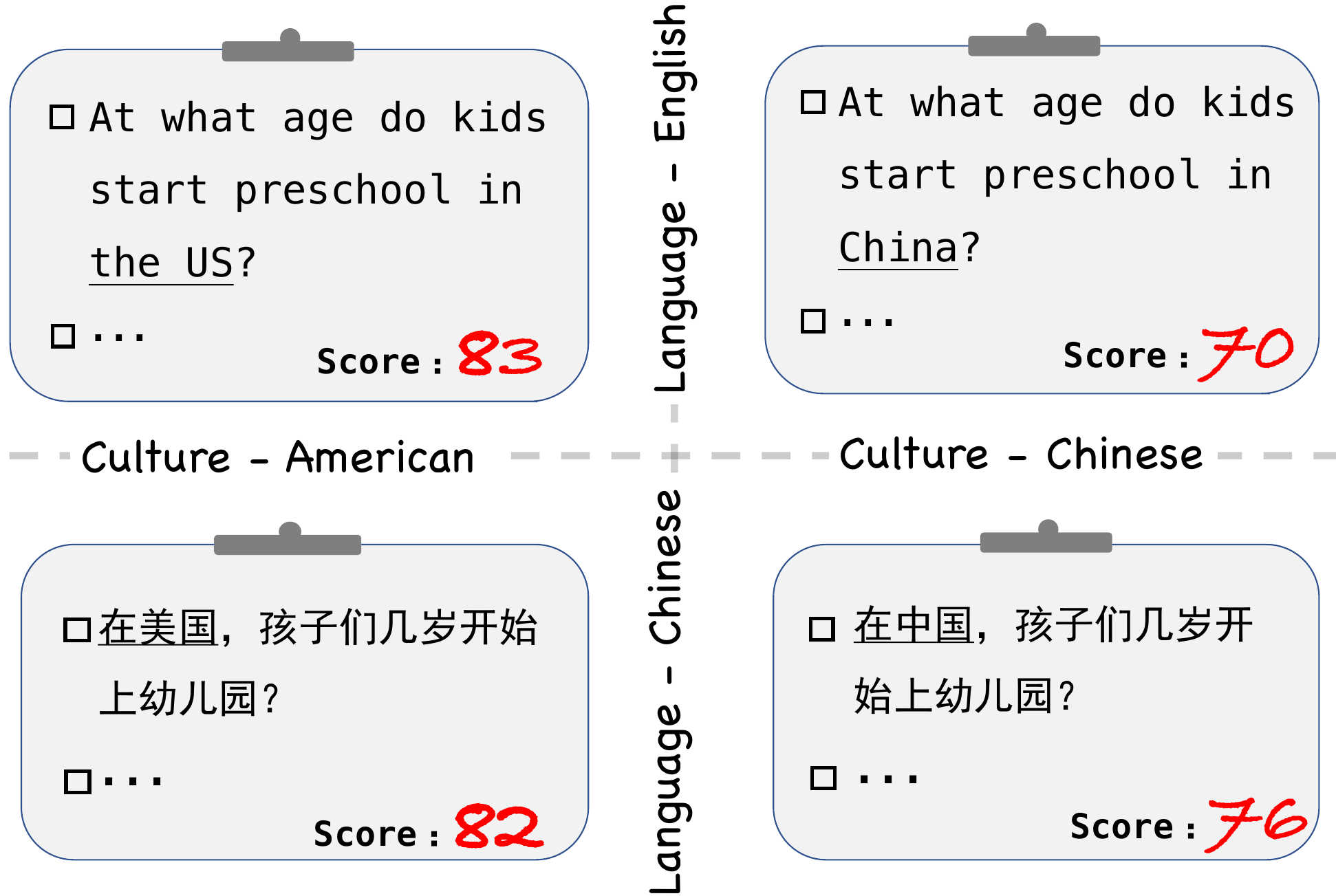}
    \caption{Dual Evaluation Framework for evaluating multilingual capabilities of LLMs. The figure is divided into four quadrants, each showing the model’s performance on questions framed in different \textbf{languages} (horizontal-axis) and \textbf{cultural contexts} (vertical-axis). The score refers to the aggregated performance of the model Claude-3.5-Sonnet on these four question sets.
    }
    \label{fig: intro}  
\end{figure}

To evaluate the LLMs' multilingual capabilities, researchers primarily rely on translating English-centric benchmarks into target languages, such as translating MMLU~\cite{hendrycks2021measuring} into MMMLU~\cite{mmmlu}. While this approach allows for efficient cross-lingual comparisons, it limits the evaluation to scenarios rooted in English-speaking cultural contexts, as the original data was predominantly collected from perspectives prevalent in English-speaking countries.
In contrast, recent work has developed culture-specific benchmarks such as M3Exam~\cite{zhang2023mexam} and BLEnD~\cite{NEURIPS2024_8eb88844}, where evaluation data are sourced from authentic, real-world scenarios in native-speaking regions.
While these better capture the majority of local usage, they also overlook that multilingual users frequently ask questions across cultural boundaries. For example, a Spanish speaker might inquire about Chinese tea usage in Spanish, while a user from China could seek details about Diwali celebrations in Chinese. 
These existing evaluations on multilingual capabilities, however, treat language and cultural context as inseparable dimensions, restricting analyses to single-language scenarios. 

To comprehensively evaluate multilingual capability, especially considering the real-world usage, we propose a Dual Evaluation framework in this paper, which decomposes the multilingual capability evaluation along two critical dimensions: (1) \textbf{linguistic medium} (the language used for questioning) and (2) \textbf{cultural context} (the regional and cultural knowledge being tested).
As illustrated in Figure~\ref{fig: intro} through a preschool enrollment example, this framework generates four distinct evaluation scenarios from a single question template.
This structured decomposition enables multiple essential multilingual capability assessments, including native cultural-linguistic alignment (same language and culture), cross-lingual understanding (different language, same culture), and cross-cultural ability (same language, different culture).

With such a dual evaluation framework design, we construct a dataset by adopting and extending the BLEnD dataset~\cite{NEURIPS2024_8eb88844}, which contains every-day questions across different cultural contexts. We then evaluate a wide range of open-source and close-source models with this newly constructed benchmark. Our findings indicate that: 1) Models generally perform better on scenarios rooted in English-speaking culture, a pattern that persists cross-lingually (Section~\ref{sec: finding1}), 
and 2) LLMs perform better when questions are posed in the language that corresponds to the cultural context of the question, rather than in English (Section~\ref{sec: finding2}). The second finding, in particular, draws our attention because most existing models are primarily trained on English data and have demonstrated strong performance in other multilingual evaluations like MMMLU. However, when faced with real-world culturally relevant questions in the corresponding language, these models perform better in that language than in English. We refer to this phenomenon as ``\textbf{Cultural-Linguistic Synergy}'' (as shown in Figure~\ref{fig: intro}, Claude-3.5-Sonnet has better performance on the Chinese test than the English test when asking about Chinese culture questions, vice versa).

To understand the underlying causes of this phenomenon, we conduct interpretability probing by analyzing the activation status of neurons when answering questions in different languages and cultural contexts, we find that: 1) The proportion of specific neurons tends to be higher when the question is in the corresponding language and cultural context, which could explain the observed ``Cultural-Linguistic Synergy'' (Section~\ref{sec: hypo1}); 
2) Additionally, this proportion of specific neurons could serve as a potential indicator for comparing multilingual capabilities during model training (Section~\ref{sec: hypo2}); 3) The number of neurons activated in the model is strongly correlated with the model’s performance in the corresponding language. Specifically, when the question is in the English-speaking cultural context, the model tends to activate more neurons, leading to better performance (Section~\ref{sec: hypo3}).

Our main contributions can be summarized as:
\setlist[itemize]{left=0pt}
\begin{itemize}
\item We propose a Dual Evaluation Framework, which decomposes the multilingual capability evaluation along two critical dimensions, linguistic medium and cultural context.
\item Through extensive experiments, we find the Cultural-Linguistic Synergy phenomenon: the selected models perform better on native cultural scenario questions when asked in the corresponding language, compared to English.
\item We demonstrate that the proportion of specific neurons activated for a given language can explain the observed Cultural-Linguistic Synergy, and that this proportion can serve as a potential indicator for comparing multilingual capabilities.
\end{itemize}

\section{Dual Evaluation Framework} ~\label{sec: dual evalution framework}

To comprehensively assess the multilingual capabilities of LLMs, we propose a Dual Evaluation framework that evaluates along two critical dimensions: (1) \textbf{linguistic medium} (the language used to pose questions) and (2) \textbf{cultural context} (the regional and cultural knowledge being tested). 
This dual-axis approach reflects three fundamental requirements for real-world applications: first, the ability to handle native language queries within their cultural context (e.g., answering ``\textit{What is a common children's snack in Spain?}'' in Spanish), second, the capability for cross-lingual understanding,  (e.g., answering questions about Spanish culture in Spanish and English); and third the capability to address cross-cultural inquiries through a single linguistic medium (e.g., answering ``\textit{What is a traditional festival in Japan?}'' in English). By evaluating LLMs in both dimensions, we can measure how well models adapt to language-specific usage scenarios while maintaining cross-lingual and cross-cultural competence.

Formally, we represent the evaluation question as $Q_{i,j}$, where $i$ denotes the cultural context and $j$ specifies the linguistic medium of the question. To construct a question set, we conduct a systematic adaptation of a culture-specific benchmark, BLEnD~\cite{NEURIPS2024_8eb88844}, testing everyday knowledge across diverse cultures and languages. Specifically, for native cultural-linguistic pairs (i.e., $Q_{i,i}$), we used the localized questions in BLEnD, which are constructed based on a template question with three key modifications: replacing country or region references, adapting phrasing to match linguistic norms, and curating culture-specific answer sets. The localized evaluation sets $Q_{i,i}$ for language $i$ are denoted by: 
\begin{equation}
\resizebox{0.89\hsize}{!}{$
    Q_{i,i} = \{(q_i, a_i) \vert (q_i, a_i) = Adapt_{i}(q),  q \in \text{Template}\}, 
$}
\end{equation}
where $Adapt_{i}$ represents the localized modifications for language $i$, and $q$ represents the template question from the $\text{Template}$ set in BLEnD.
For example, by adapting the original question \textit{``What is the most popular sports team in your country?''} into \textit{``What is the most popular sports team in the US?''}, where $i=en$, we can test the model's ability to handle English in the US context. Using these adapted questions for different languages, we can assess the model’s ability to handle native-language queries within various cultural environments.

On the other hand, to assess the model’s ability to handle questions across multiple cultural contexts when asked in a single language, we extend the $Q_{i,i}$ sets into localized transformations $Q_{i,j}$ for each language pair $(i, j)$, where $i \neq j$. 
The original BLEnD includes, for each language-specific evaluation set $Q_{i,j}$ (except for English), an English translation evaluation dataset $Q_{i,en}$. Specifically:
\begin{equation}
\resizebox{0.89\hsize}{!}{$
Q_{i,en} = \left\{ \text{Trans}_{en}(q_{i}, a_{i}) \mid  (q_i, a_i) \in Q_{i,i}, i \neq \text{en} \right\}.
$}
\end{equation}
For other language pairs $(i, j)$, we use the GPT-4o model, known for its strong multilingual capabilities, to construct these cross-lingual datasets. 

\begin{equation}
\resizebox{0.89\hsize}{!}{$
Q_{i,j} = \left\{ \text{Trans}_{j}(q_{i}, a_{i}) \mid  (q_i, a_i) \in Q_{\text{i}}, j \neq \text{en} \right\}.
$}
\end{equation}
This setup enables assessing how well the model can adapt to answering questions posed in one language about the cultural context of another.

Combining the two evaluation scenarios, the complete evaluation set $Q$ is thus represented as:
\begin{equation}
\resizebox{0.70\hsize}{!}{$
Q = \bigcup_{i \neq j} Q_{i,i} \cup Q_{j,j} \cup Q_{i, j} \cup Q_{j,i}.
$}
\end{equation}
This Dual Evaluation framework, where questions are tailored to the linguistic medium and the corresponding cultural contexts of usage, not only assesses LLMs’ multilingual abilities from both native usage scenarios ($Q_{i,i}$) and cross-cultural contexts ($Q_{j,i}$) but also employs a completely dual-format question approach.  
Specifically, $Q_{i,i}$ and $Q_{j,j}$ are constructed using the same template question, and tailored to different linguistic and cultural contexts. This approach allows us to quantitatively compare the multilingual capabilities cross-culturally within the same language (by comparing $Q_{i,i}$ and $Q_{j,i}$), and cross-lingually (by comparing $Q_{i,i}$ with $Q_{j,j}$, or $Q_{i,i}$ with $Q_{i,j}$). An example of this dual evaluation sample is shown in Figure~\ref{fig: intro}, and the details of the completion for $Q_{i,j}$ and human evaluation for the quality are presented in Appendix~\ref{appen: completion for Dataset}.

\begin{table}[tb]
\centering
 \resizebox{0.48\textwidth}{!}{%
\begin{tabular}{llcccr}
\toprule
\textbf{Linguistic Medium} & \textbf{Cultural Context} & \textbf{\# Data Sample} \\
\midrule
English (en) & \textbf{United States (US)}, CN, ES, ID, & 3,500\\
             & KR, IR, JB  \\
Chinese (zh)    & \textbf{China (CN)}, \textcolor{green}{\underline{US}}    & 1,000\\ 
Spanish (es)    & \textbf{Spain (ES)}, \textcolor{green}{\underline{US}}      & 1,000\\
Indonesian (id) & \textbf{Indonesia (ID)}, \textcolor{green}{\underline{US}}  & 1,000\\
Korean (ko)     & \textbf{South Korea (KR)}, \textcolor{green}{\underline{US}} & 1,000\\
Persian (fa)    & \textbf{Iran (IR)}, \textcolor{green}{\underline{US}}       & 1,000\\
Sundanese (su)  & \textbf{West Java (JB)}, \textcolor{green}{\underline{US}}  & 1,000\\
\midrule
\textbf{Total} &  &9,500\\
\bottomrule
\end{tabular}}
\caption{Overview of the evaluation dataset, detailing the language, cultural context of certain countries/regions, and the number of data samples. 
\textbf{Bolded} countries/regions indicate where the corresponding language is spoken natively,
while the others are transformed for cross-cultural evaluation. Each language has 500 data samples per country/region. The parts we added to the original BLEnD are marked in \textcolor{green}{\underline{green}}.}
\label{tab: dataset static}
\end{table}

\section{Multilingual Capabilities Evaluation}

\subsection{Experiment setting}
 We select a wide range of LLMs of different sizes to evaluate their multilingual capabilities, including GPT-4o~\cite{openai2024gpt4technicalreport}, Claude-3.5-Sonnet~\cite{claude35}, CommandR~\cite{commandR}, Llama-3-8B-Instruction, Llama-3-70B-Instruction\cite{grattafiori2024llama3herdmodels}, Gemma-2-9B~\cite{gemmateam2024gemma2improvingopen},  Qwen2.5-7B-Instruct~\cite{qwen2025qwen25technicalreport}, and Bloomz-7B~\cite{muennighoff2022crosslingual}. 
 The experiment is conducted across seven languages, with cultural content sourced from one of the typical countries where each language is widely spoken. 
Considering the current performance of the model (primarily strong in English), and taking cost and time constraints into account, we only construct evaluation data $Q_{i,j}$ for the language pairs $(i, j)$ where either language $i$ or $j$ is English.
The language and culture information for the evaluation dataset are provided in Table~\ref{tab: dataset static}. 
The questions used in evaluation are short-answer questions (SAQs) aligned with the BLEnD~\cite{NEURIPS2024_8eb88844} benchmark. We apply non-weighted scores for the evaluation metrics. During inference, all models are tested in a zero-shot setting. The question prompts are derived from the original BLEnD instruction set. More details are shown in Appendix~\ref{appen: instruction}.

\subsection{Finding 1: LLMs' Performance Declines as the Cultural Context Shifts from English to Cross-Cultural Scenarios.} \label{sec: finding1}

\begin{figure}[htb]
    \centering
    \includegraphics[scale=0.32]{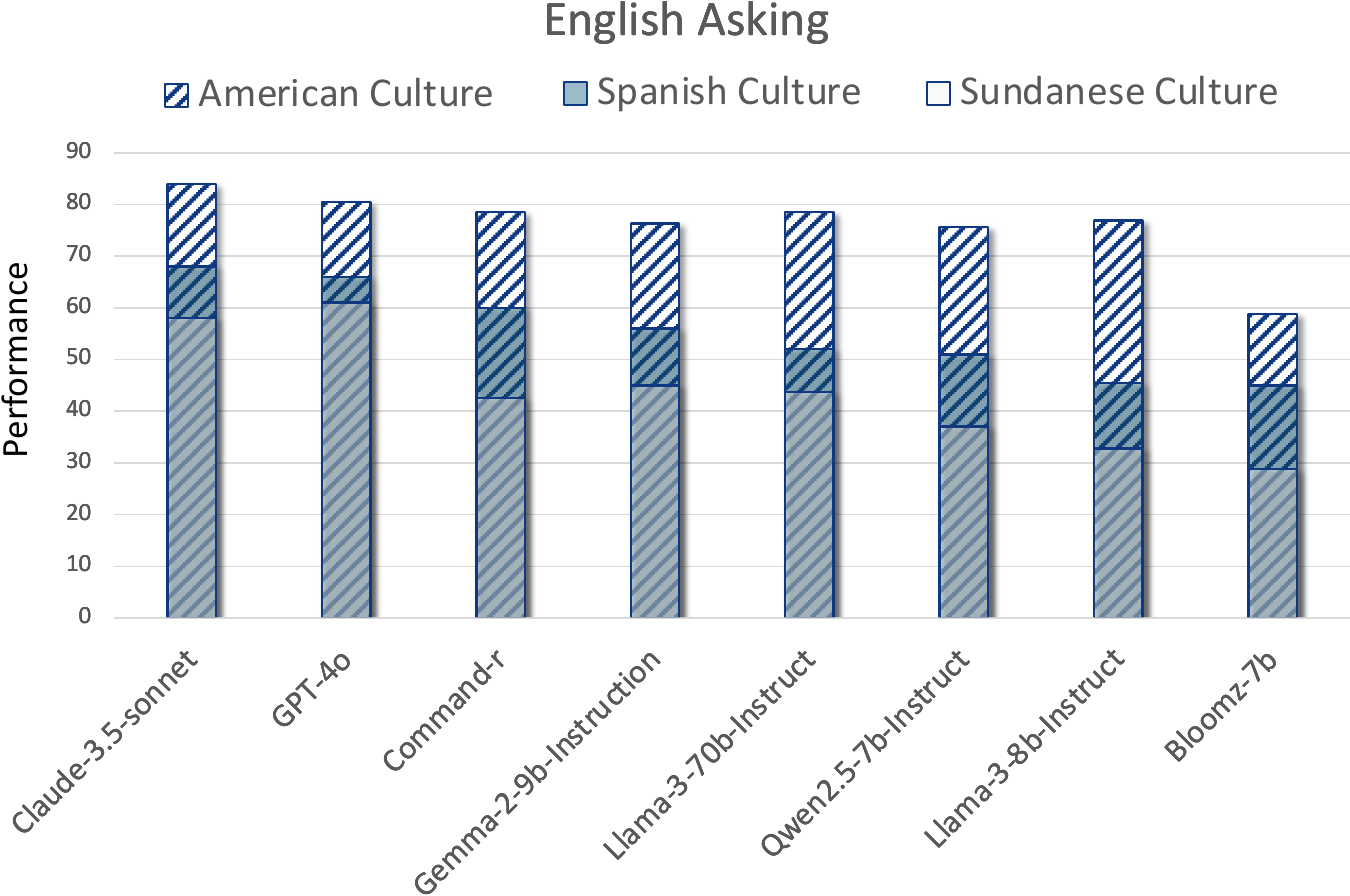}
    \caption{The performance of the selected models on American, Spanish, and Sundanese culture questions when asked in English. We find that models perform best on American culture. }
    \label{fig: finding1}
\end{figure}

Using our Dual Evaluation Framework, we evaluate the performance of the selected LLMs. As mentioned in Section~\ref{sec: dual evalution framework}, one of the key advantages of this framework is its ability to compare models’ adaptability in cross-cultural contexts (comparing $Q_{i,i}$ \& $Q_{j,i}$), given that the questions are presented in a dual format. 
Since most of the selected models primarily use English as their training corpus, we first compare the performance on $Q_{en,en}$ and $Q_{i,en}$ (where $i \neq \text{en}$). 
The results in Figure~\ref{fig: finding1} (full result in Append~\ref{append: more results}), for each bar, represent the performance in specific culture-contexts. By comparing the bars' height across different context, we observe that models perform best for English-speaking culture contexts, when asking questions in English, and performance declines in other language-speaking cultural context, with the drop becoming more pronounced as the language’s resource availability decreases.

While this trend echoes previous findings with translated datasets from English culture~\cite{NEURIPS2024_8eb88844, mmmlu}, it raises a further question: Does this phenomenon also hold in other languages?
\begin{table}[t]
\centering
\resizebox{0.46\textwidth}{!}{%
\begin{tabular}{l|c|c}
\toprule
\textbf{Question Content} & \textbf{Spanish Culture} & \textbf{American Culture} \\
\midrule
Claude-3.5-Sonnet & 81.0 & \textbf{82.0} \\
GPT-4o & 76.5 & \textbf{77.6} \\
Command-r & 69.9 & \textbf{73.4} \\
Gemma-2-9b-Instruction & 70.9 & \textbf{72.7} \\
Llama-3-70b-Instruct & 72.0 & \textbf{79.6} \\
Qwen2.5-7b-Instruct & 62.0 & \textbf{70.5} \\
Llama-3-8b-Instruct & 58.9 & \textbf{74.5} \\
Bloomz-7b & \textbf{53.6} & 52.8 \\
\bottomrule
\end{tabular}
}
\caption{The performance of selected models on everyday questions about Spain and the US when asked in \textbf{Spanish}. Generally, models perform best when asking questions about US culture in Spanish.}
 \label{tab: finding1_1}
\end{table}
To explore more, we expanded the comparison to include $Q_{i,i}$ and $Q_{i, en}$, especially when $i \neq \text{en}$. The results for $i=\text{es}$ (Spanish) are shown in Table~\ref{tab: finding1_1}, considering the high availability of resources of it.
Additional results, demonstrating the same phenomena, are available in Appendix~\ref{append: more results}. 
The results indicate that, in general, the selected models perform better on English-speaking culture questions compared to other languages when asked in the respective language. Since the training data for these models is not fully open-source, we hypothesize that, for each language in the training corpus, the models are trained on a larger volume of English-language usage scenarios. As a result, the models exhibit better performance on English-speaking culture questions across all languages. In the following interpretability Section~\ref{sec: hypo3}, we delve deeper into the model’s internal workings to explore the reasons behind this observed behavior.

\subsection{Finding 2: LLMs perform better when asked in the corresponding language.} \label{sec: finding2}
\begin{figure*}[htp]
    \centering   
    \includegraphics[scale=0.3]{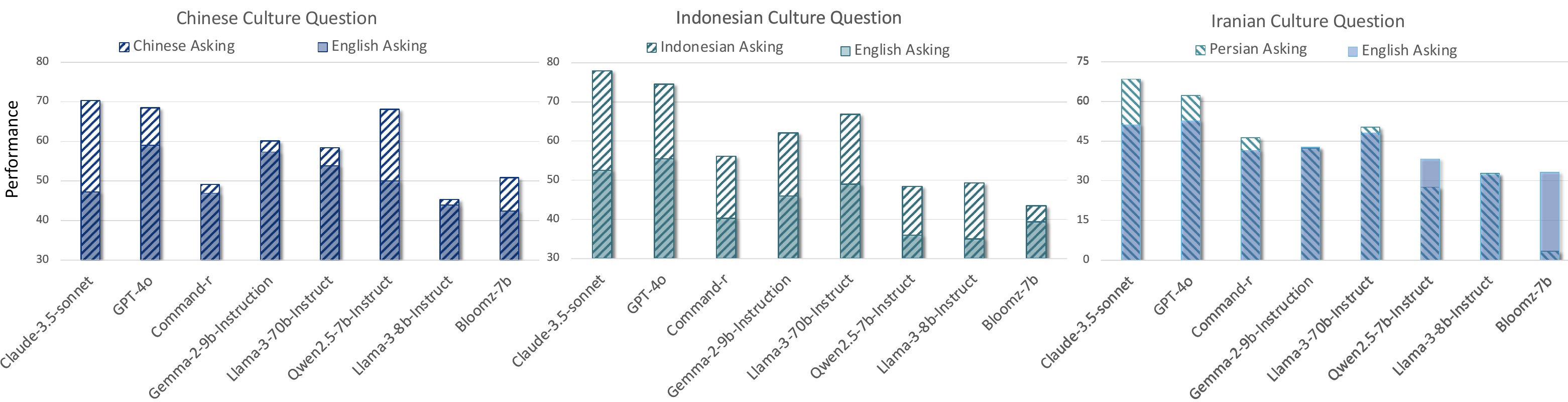}
    \caption{The performance of the selected models on Chinese, Indonesian, and Iranian culture questions when asked in the corresponding language versus English. We generally find that models perform better when the questions are posed in the corresponding major spoken language compared to English.}
    \label{fig: finding2}
\end{figure*}

In addition to enabling comparisons of behavior in cross-cultural contexts within the same language (as discussed in Section~\ref{sec: finding1}), we can evaluate how models perform cross-lingually using Dual Evaluation Framework. Specifically, by comparing $Q_{i,i}$ and $Q_{i,j}$ ($i \neq j$), we get the result shown in Figure~\ref{fig: finding2}. We surprisingly find that asking culture-related questions in the corresponding language outperforms asking in English, as indicated by the bars with patterns being higher than those without.
Specifically, across the eight selected models, the average performance for questions related to Chinese culture when asked in Chinese exceeded that of asking in English by 8.8 points. Similarly, questions related to Indonesian culture posed in Indonesian outperformed those asked in English by 15.7 points.
While this advantage diminishes when dealing with lower-resource languages. For instance, in Persian, the performance gap is  -0.95 points. This can be attributed to models like Bloomz-7B, which have limited or no training data for Persian, resulting in better performance when asking questions in English instead. On the other hand, this corresponding advantage also appears in American culture questions, as shown in Figure~\ref{fig: finding2_1}.

From these observations, we can generally summarize that asking culture-specific questions in their corresponding language tends to outperform answering them in English. 
We refer to this counterintuitive phenomenon as  \textbf{``Cultural-Linguistic Synergy''}. 
That is, aligning the cultural context with the appropriate linguistic medium, we can achieve superior performance --- even for models primarily trained on English data, which perform better on English-specific tasks than on other language benchmarks like MMMLU and translated GSM8K
~\cite{shi2022languagemodelsmultilingualchainofthought}.
An intuitive explanation for this Cultural-Linguistic Synergy could lie in the training data. However, due to the lack of access to the training data and the massive scale of the training corpus, further exploration in this direction can be challenging.
Thus, in the following sections, we proceed with interpretability analysis to understand the mechanisms of this Cultural-Linguistic Synergy, beginning with preliminary insights.

\section{Interpreting Cultural-Linguistic Synergy}
\subsection{Preliminary}

\begin{figure*}[htp]
    \centering   
    \includegraphics[scale=0.26]{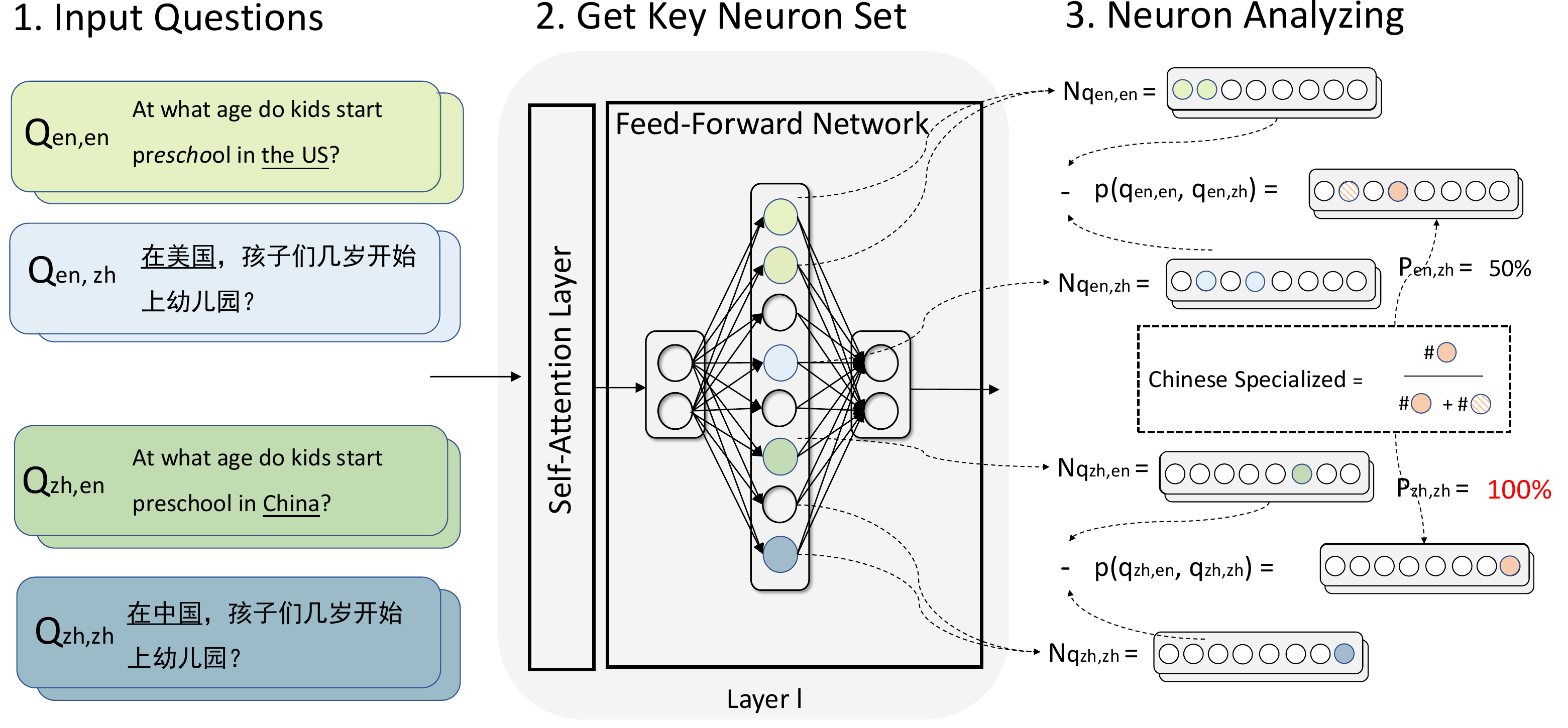}
    \caption{Workflow for interpreting Cultural-Linguistic Synergy. For every dual-format question pair, we trace neuron activations and measure the share of language specialized neurons (Chinese in the above figure) that fire when each question is posed.
    }
    \label{fig: probing}
\vspace{-4mm}
\end{figure*}

\noindent \textbf{Neurons in FFN Module:}
Recent interpretability studies suggest that factual knowledge is stored in the FFN memories and represented by neurons in the network \citep{geva2021transformer}. Given the input token $x$, the FFN module of layer $l$ in a decoder-only Transformer can be represented as (outer activation functions and bias terms are omitted for clarity):
\begin{equation}
\resizebox{0.89\hsize}{!}{$
\text{FFN}^{l} (h^{l})= \left(W_{down}^{l} \cdot Activation (W_{up}^{l} \cdot h^{l} ) \right)
$}
\end{equation}
where $h^{l}$ is the input to the FFN, $W_{up}^{l}$ and $W_{down}^{l} $ are the weight matrices, and $Activation$ is the activation function.
Following previous works, 
the $i$-th element of $\textstyle Activation(W_{up}^{l} \cdot h^{l} ) \in \mathbb{R}^{dm}$ is considered the $i$-th neuron in layer $l$ (a simple illustration of neuron in Figure~\ref{fig: probing}). The value of this neuron for the input token $x$ can be represented by its corresponding activation value ${v}^x_{(i,l)}$.

\noindent \textbf{Key Neuron Set: }
Following previous work, neurons with higher activation values when answering a question are considered more important \citep{tang2024language, zhao2024large, hong2024intrinsicevaluationunlearningusing, cao2025modelutilitylawevaluating}. Therefore, given a question $q$ , we can identify the ``Key Neurons'' $N_q$ by selecting neurons that are highly activated in the model’s response $r = \{r_1, r_2, \dots, r_n\}$, where $r_i$ denotes the $i$-th token in the response, based on a threshold function ($threshold$) as:
\begin{equation}
\resizebox{0.75\hsize}{!}{$
N_{q} = \left\{ (i, l) \mid \text{v}_{(i, l)}^{r_i} > threshold, r_i \in r \right\}
$}
\label{equ: neruon set}
\end{equation}

\noindent By aggregating these key neurons for each question $q$ in the dataset $Q = \{q\}$, we obtain the Key Neuron set for the entire dataset $Q$ as (ref Figure~\ref{fig: probing} for illustration for getting key neurons):
\begin{equation}
\resizebox{0.65\hsize}{!}{$
N_{Q} = \left\{ (i,l) \vert (i,l) \in N_{q}, q \in Q  \right\}
$}
\end{equation}

\subsection{Experiment Setup} \label{sec: setup}
Considering that the Cultural-Linguistic Synergy arises from variations in cultural context, we investigate how the model’s internal behavior differs when asking questions in language $i$. Specifically, we focus on two types of question content: one related to the American cultural context and the other to the cultural context of language $i$. By comparing these two contexts, we aim to uncover why, in the latter case, asking questions in the corresponding language leads to better performance than asking them in English.
To explore this, we focus on calculating the ``specialized neurons'' activated in each context.  These specialized neurons refer to the Key Neurons that activate when answering in language $i$ , as opposed to English.
For American cultural context, we obtain the Key Neuron sets $N_{q_{en, en}}$ and $N_{q_{en, i}}$ for the dual-ed questions $q_{en, en}$ and $q_{en, i}$ from $Q_{en, en}$ and $Q_{en, i}$, respectively. By gathering the neuron only activate when asking $q_{en, i}$ , we can determine the proportion of specialized language $i$ neurons for the question pair $(q_{en,en}, q_{en, i})$ as:
\begin{equation}
\resizebox{0.65\hsize}{!}{$
p(q_{en,en},  q_{en, i}) = \frac{ \left| N_{q_{en, i}} - N_{q_{en,en}} \right|}{ \left| N_{q_{en, i}} \right|}
$}
\end{equation}
For example, as shown in Figure~\ref{fig: probing}, we calculate the key neurons for the paired question \textit{``At what age do kids start preschool in the US?’’} in both English and Chinese, to identify the specialized Chinese neurons (depicted in red). We then repeat for every question pair $(q_{en,en}, q_{en,i})$, and compute the average proportion of specialized neurons $p(q_{en,en}, q_{en,i})$ across all dual-ed question pairs. This gives us the proportion of specialized neurons for language $i$ in the American cultural context, denoted as $P_{en,i}$. Similarly, we calculate the proportion of specialized neurons for $i$ in the cultural context of language $i$, denoted as $P_{i,i}$.
By comparing the proportions of specialized neurons between these two contexts, we aim to find the underlying factors contributing to Cultural-Linguistic Synergy.

For time and cost efficiency considerations, we deploy Qwen2.5-7B-Instruction and Llama-3-8B-Instruction models as the target model. To obtain the Key Neuron Set, we use the instruction 2 (Appdenx~\ref{appen: instruction}) to get the response $r$ and apply the top-k ($k=5$) threshold for each layer, as defined in Equation~\ref{equ: neruon set} (more details about the threshold function is shown in Appendix~\ref{appen: thresholdselection}. The details of the selection for this hyperparameter can be found in Section~\ref{sec: abla}.

\subsection{Analyzing} \label{sec: hypo1}

We compare $P_{en,i}$ between $P_{i,i}$ ($i \neq \text{en}$) across the six languages.
As shown in Figure~\ref{fig: hypo1}, generally, $P_{i,i}$ (bars with patterns) is higher than $P_{en,i}$ (bars without patterns) in the scenarios where model demonstrates the Cultural-Linguistic Synergy (e.g., Llama-3-8B in Chinese, Indonesian, Persian, and Korean, Qwen2.5-7B in Chinese, Korean). Conversely, when no Cultural-Linguistic Synergy is observed, $P_{i,i}$ is lower than $P_{en,i}$ (e.g., Llama-3-8B in Sundanese, Qwen2.5-7B in Persian, Sundanese). 
This suggests that \textbf{models tend to activate a higher proportion of neurons specialized for the target language when the cultural context aligns with the corresponding linguistic medium, compared to when this alignment is absent.} The activation of these specialized neurons allows the model to better utilize knowledge specific to the culture and the target language.
This knowledge, which may not be fully accessed when asking in English, contributes to the model’s better performance in the target language.
However, Spanish stands as an exception, which we attribute to the high similarity between Spanish and English in terms of language structure, and thus may have greater overlap of the knowledge-storing neurons.

\subsubsection{Hypothesis 1 and Validation} \label{sec: hypo2}

\begin{figure}[htb]
    \centering
    \includegraphics[scale=0.36]{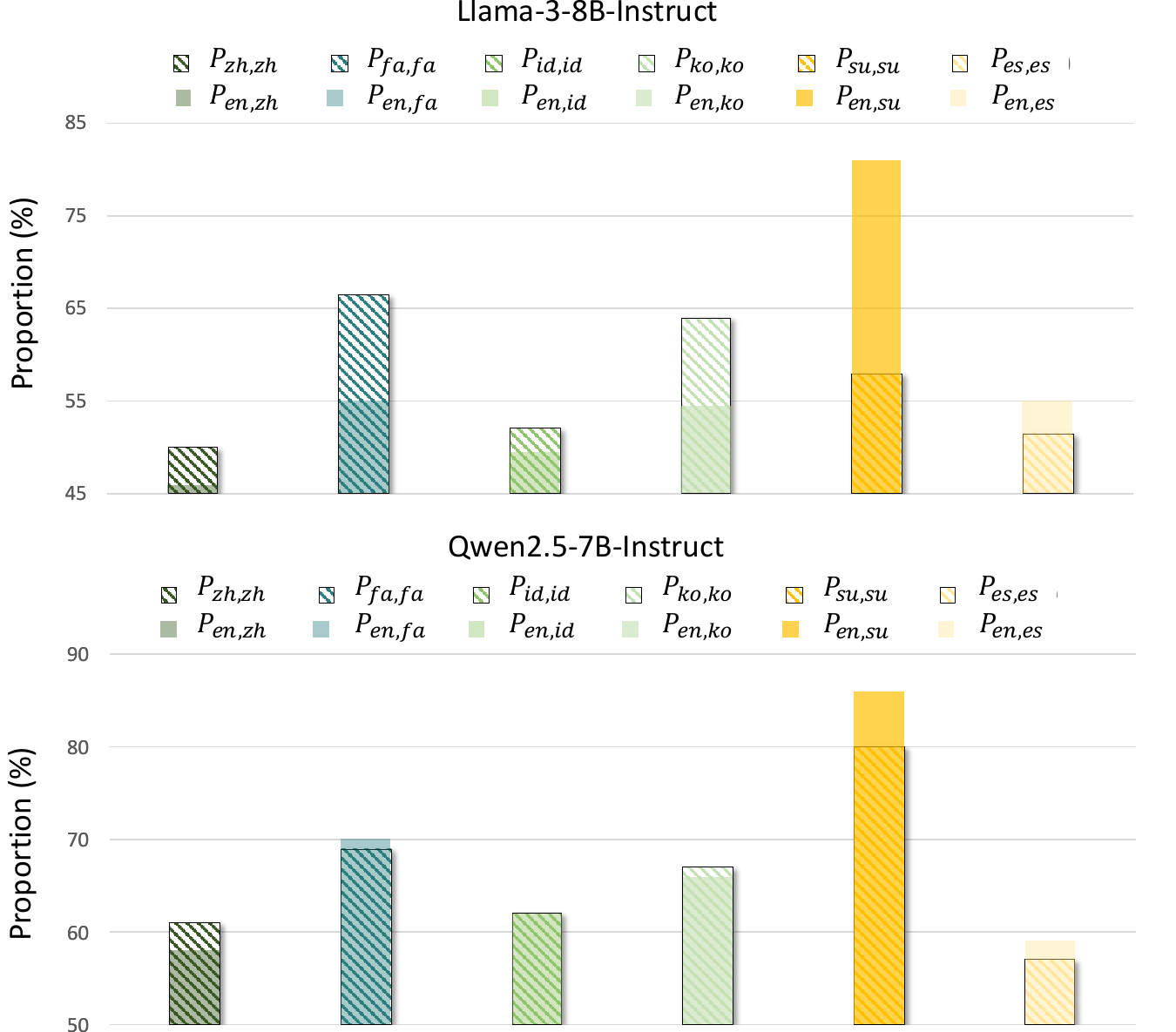}
    \caption{Comparison of the proportion of specialized neurons for language Chinese(zh),  Spanish(es), Indonesian(id), Korean(ko), Persian(fa), and Sundanese(su) between different cultural context. It indicates that when Cultural-Linguistic Synergy happens, models generally activate a higher proportion of specific neurons.}
    \label{fig: hypo1}
\end{figure}

Previous analysis (Section~\ref{sec: hypo1}) suggests that when Cultural-Linguistic Synergy occurs, the model activates a higher proportion of neurons specialized for the language and culture. This ability to better utilize knowledge aligned with the corresponding cultural context helps guide the model to perform better than when asking in English. Building on this, we further consider whether more powerful multilingual models have a better ability to utilize culture and language-specific knowledge. This could, in turn, serve as a valuable metric for evaluating model performance during training.

\begin{tcolorbox}[width=0.99\hsize,opacityfill=0.1]
\textbf{Hypothesis 1:} 
Models have a better ability to utilize cultural knowledge will activate a \textbf{higher proportion} of specialized neurons when the cultural context aligns with the linguistic medium.
\end{tcolorbox}
\noindent Figure~\ref{fig: hypo1} indicates that Qwen2.5 utilizes more specialized neurons (66 \%) than Llama-3 (57\%) across the six languages, which may provides evidence for this hypothesis. However, note that differences in training data and model architectures between different series may limit the direct comparability.

On the other hand, validating this hypothesis by improving one model with additional training data may present challenges. This is due to the limited availability of language resources and the potential risk of benchmark leakage, which could affect the analysis. Thus we use well-recognized multilingual same series models with distinct language capabilities, such as the open-source multilingual extension of the Llama-3 model, Llama-3.1-8B-Instruction, for comparative analysis with Llama-3-8B-Instruction in the validation experiment.

\begin{figure}[htb]
    \centering
    \includegraphics[scale=0.37]{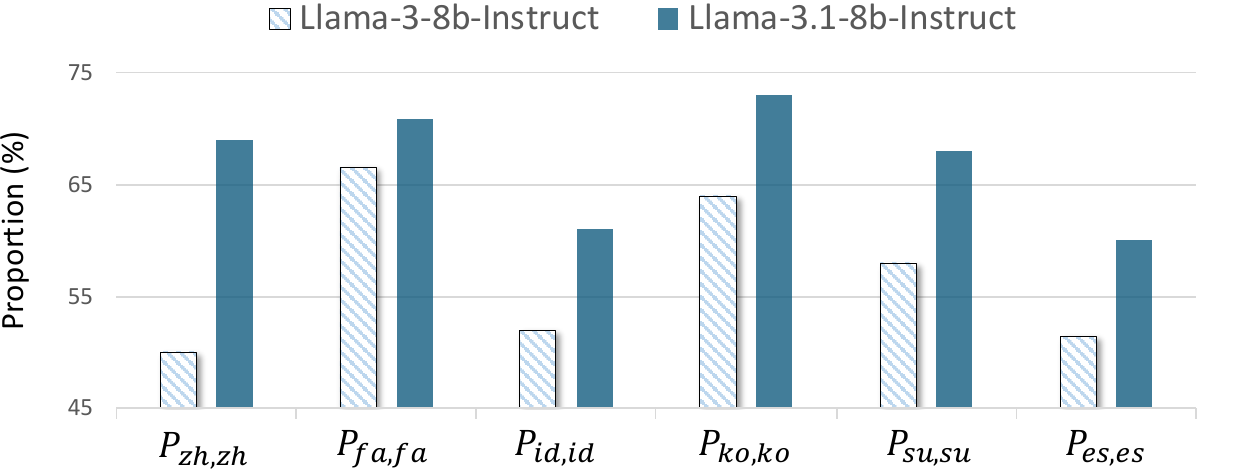}
    \caption{Comparison of proportion of specialized neurons for Llama-3-8B-instruct and Llama-3.1-8B-instruct. The result shows that Llama-3.1-8B-instruct, the multilingual extension for Llama-3-8B-instruction, has a higher proportion of specialized neurons.}
    \label{fig: hypo2}
    \vspace{-3mm}
\end{figure}

The results shown in Figure~\ref{fig: hypo2} indicate that Llama-3.1-8B-Instruction activates a higher proportion of specialized neurons (67\%) compared to Llama-3-8B-Instruction (57\%), supporting our hypothesis that models with stronger capabilities in the corresponding language are better at leveraging language-specific neurons.
Furthermore, this proportion of the specific neurons could be utilized as a potential \textbf{indicator for evaluating a model’s ability to effectively leverage multilingual knowledge} during the training phase.

\subsubsection{Hypothesis 2 and Validation}  \label{sec: hypo3}

Through previous analysis (Section~\ref{sec: hypo1}), we find that the proportion of specific neurons may be indicative of the Cultural-Linguistic Synergy. 
It left us thinking: If a higher proportion of neurons corresponds to greater knowledge neuron utilization by the model, then assuming a consistent increase in the proportion of language-specific neurons for one specific model, we expect that an increase in the number of neurons should lead to better performance for the corresponding language. 

\begin{tcolorbox}[width=0.99\hsize,opacityfill=0.1]
\textbf{Hypothesis 2:}
The greater the \textbf{number of neurons} activated for questions in a given language, the better the performance.
\end{tcolorbox}

\noindent Since there is no consensus on how to definitively measure the importance of individual neurons, we take a different perspective. Instead of focusing on neuron quantity directly, we explore whether the total number of neurons activated across the dataset is correlated with the model’s performance.

\begin{figure}[htb]
    \centering
    \includegraphics[scale=0.32]{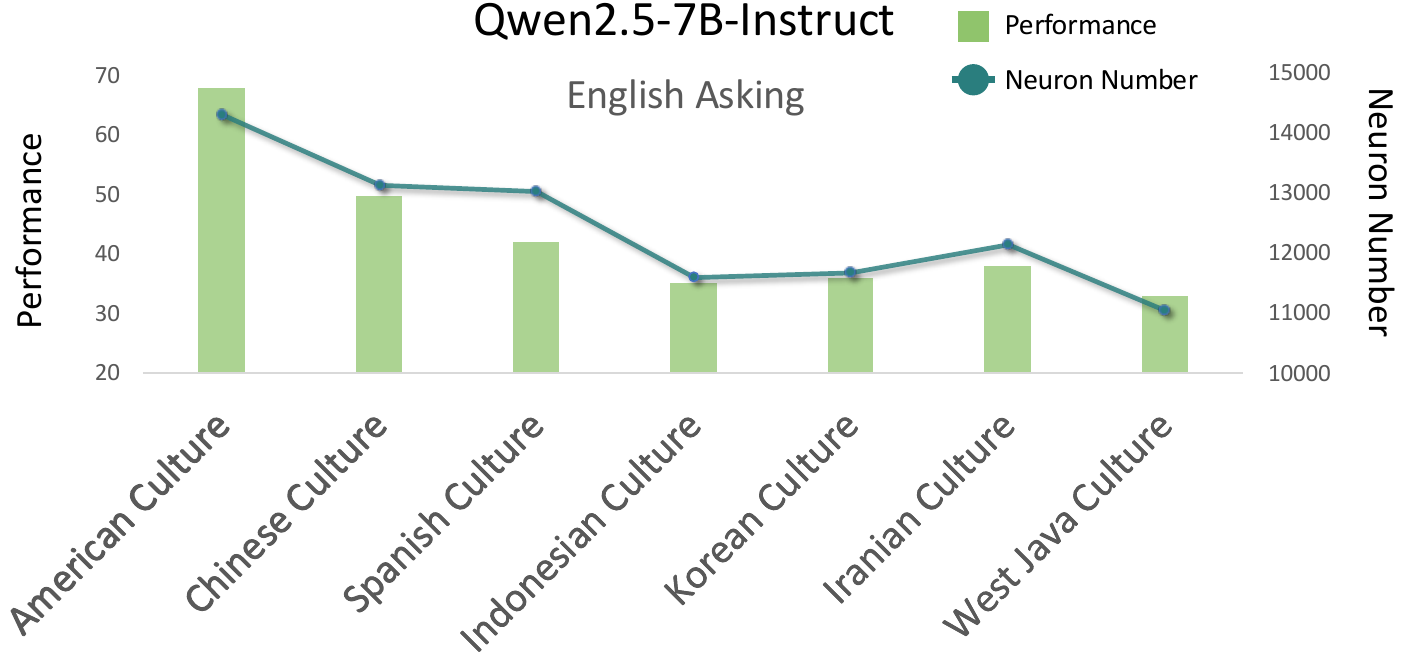}
    \caption{The performance and the number of Key Neurons for the Llama-3-8B on cross-cultural contexts.}
    \label{fig: hypo3}
    \vspace{-3mm}
\end{figure}

From Section~\ref{sec: hypo1}, we notice that knowledge representation may vary across languages. Therefore, in this validation study, we focus on comparisons within the same language. Specifically, we investigate the relationship between the set of Key Neurons set, $|N_{Q_{i,i}}|$ and $|N_{Q_{j, i}}|$, and the model’s performance on the corresponding evaluation data. The results, shown in Figure~\ref{fig: hypo3}, indicate that the total number of activated neurons is highly correlated with the model’s performance, with a Pearson correlation coefficient of 0.95 for English questions (more results are in Appendix~\ref{append: more results}). This suggests that when more neurons are activated, the model is likely utilizing more relevant knowledge, leading to better performance. This finding aligns with the observation in Section~\ref{sec: finding1}, where, in American cultural context, the model activates the most neurons.

\section{Ablation Study} \label{sec: abla}

\begin{figure}[tb]
    \centering
    \includegraphics[scale=0.33]{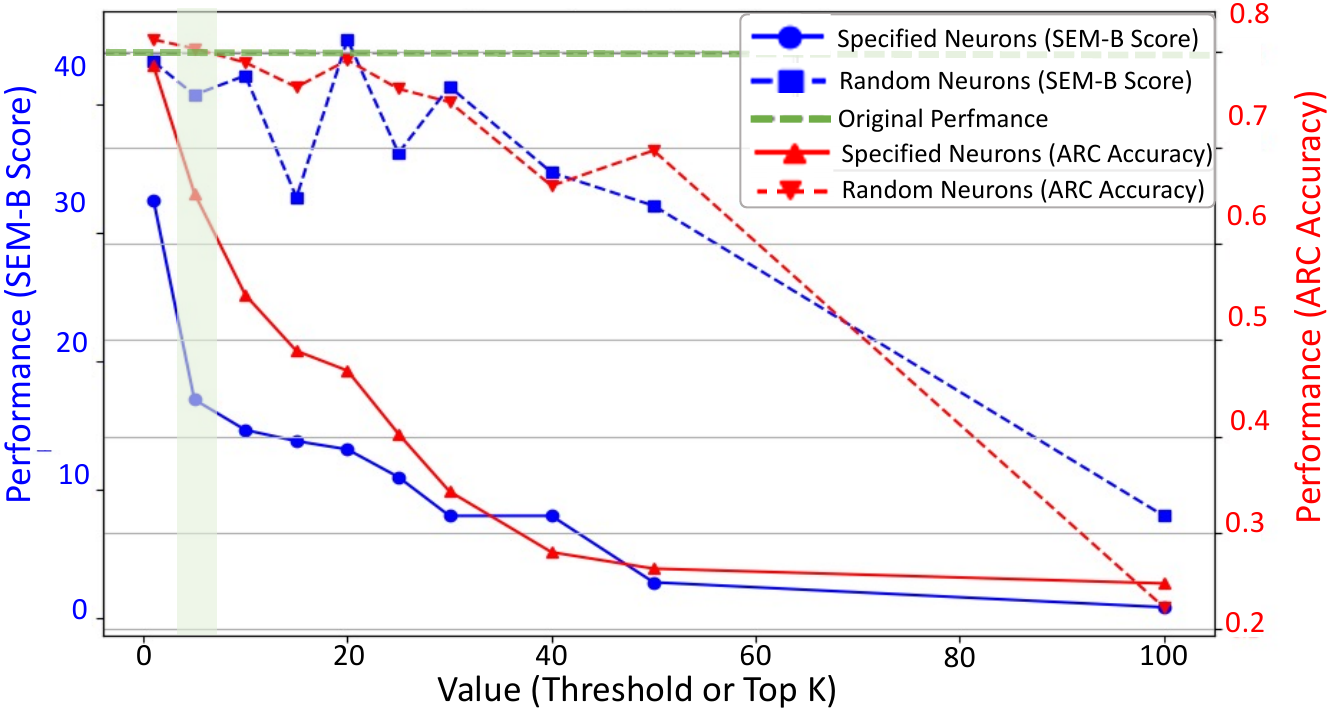}
    \caption{Performance for Llama-3-8B-Instruction on ARC and $Q_{en,en}$ when masking Key Neurons with different threshold.}
    \label{fig: abli}
    \vspace{-3mm}
\end{figure}
In our experimental setup (Section~\ref{sec: setup}), we select $k = 5$ as the threshold. The threshold is set to ensure that the selected key neurons accurately represent the model’s knowledge on the given question.  To determine the optimal threshold, we measure the performance drop when masking the corresponding neurons on the selected task. We choose the threshold where the performance drops significantly for the masked neurons, while the performance on out-of-distribution (OOD) knowledge (here we use the ARC~\cite{allenai:arc} dataset), is largely unaffected. As shown in Figure~\ref{fig: abli}, we draw in blue and red, respectively. We also use a random mask with the same number of neurons as a baseline (dash line). The selected threshold is depicted in green.
\section{Related Work}
\textbf{Multilingual Capabilities Evaluation}
To evaluate the LLMs' multilingual capabilities, researchers use translating English-centric benchmarks such as  MMMLU~\cite{mmmlu}, MGSM~\cite{shi2022languagemodelsmultilingualchainofthought} and Multilingual MT-Bench~\cite{zheng2023judging}. Recent work also developed culture-specific benchmarks~\cite{cao2025generalizableevaluationllmera, zhang2023mexam, NEURIPS2024_8eb88844, leong2023bhasaholisticsoutheastasian, liu2025seaexamseabenchbenchmarkingllms}. For example, M3Exam~\cite{zhang2023mexam} sourced from real and official human exam questions, BLEnD~\cite{NEURIPS2024_8eb88844} where evaluation data are crafted from real-world scenarios and CulturalBench~\cite{chiu2024culturalbenchrobustdiversechallenging} with human-written questions covering 45 global regions. These existing evaluations, however, treat language and cultural context as inseparable dimensions, restricting analyses to single-language scenarios. 

\noindent \textbf{Multilingual Capabilities Interpretation}
Recently, some work~\cite{wang2024large, kojima2024multilingual, wang2024sharingmattersanalysingneurons} use the Mechanistic interpretability to analyze the model's multilingual capabilities. ~\citet{tang2024languagespecificneuronskeymultilingual} shows that proficiency in processing a particular language is predominantly due to a small subset of neurons. ~\citet{wendler2024llamasworkenglishlatent} projects the hidden state into vocabulary to investigate the Latent Language. ~\citet{zhao2024large} further proposed the multilingual workflow to understand how LLMs Handle Multilingualism. However, these studies do not investigate the model's behavior across different cultural contexts and languages.

\section{Conclusion}

This study introduced a Dual Evaluation Framework specifically designed to comprehensively assess LLMs across linguistic medium and cultural contexts. Our findings reveal ``Cultural-Linguistic Synergy,'' phenomenon where models perform optimally when questions are culturally aligned with the language, challenging the prevailing assumption that LLMs, primarily trained on English data, perform uniformly across different languages. Utilizing interpretative methods, we delved deeper into this phenomenon and found that it is related to the Key Neurons. As the field of interpretability in AI continues to evolve, we plan to further expand this framework to enable more comprehensive and nuanced evaluations of multilingual models.
\section{Limitation}

While the Dual Evaluation Framework is flexible enough to incorporate additional benchmarks, the prerequisite for conducting meaningful cross-cultural comparisons, especially to conduct neuron probing, lies in having dual-format question content. This content needs to capture both linguistic and cultural nuances. Without this dual-format structure, performing robust and quantitative cross-cultural comparisons remains limited.

In our current experimental design, we focus on a single cultural context for each language, based on typical countries or regions where the language is spoken. However, given the widespread usage of some languages, especially in regions with diverse cultural contexts, we plan to expand the framework in the future to incorporate more varied cultural contexts to make our conclusions more robust.

Due to time and computational cost constraints, we limited our probing validation to models like Qwen2.5-7B-Instruction and Llama-3-8B-Instruction. As LLMs interpretation techniques continue to evolve and improve, we plan to expand the range of models included in future studies, especially larger models with more parameters, to gain deeper insights into multilingual and cross-cultural model behavior.

\section{Acknowledgement}
This research is supported by the Ministry of Education, Singapore, under its Academic Research Fund (AcRF) Tier 1 grant, and funded through the SUTD Assistant Professorship Scheme (SAP 2025\_001).
We would also like to thank Yizhe Yang from Beijing Institute of Technology for helping us with human evaluation.

\bibliography{anthology,custom}
\clearpage
\appendix
\section{More Experiment Detail}
\subsection{Response Generation Setting}

Answer generation across the involved models is conducted in a zero-shot setting, with all models set to a temperature of 0.0 and a maximum token length of 1024.

\subsection{Result Display Setting} \label{append: more results}

\begin{figure*}[htb]
    \centering
    
    \includegraphics[scale=0.42]{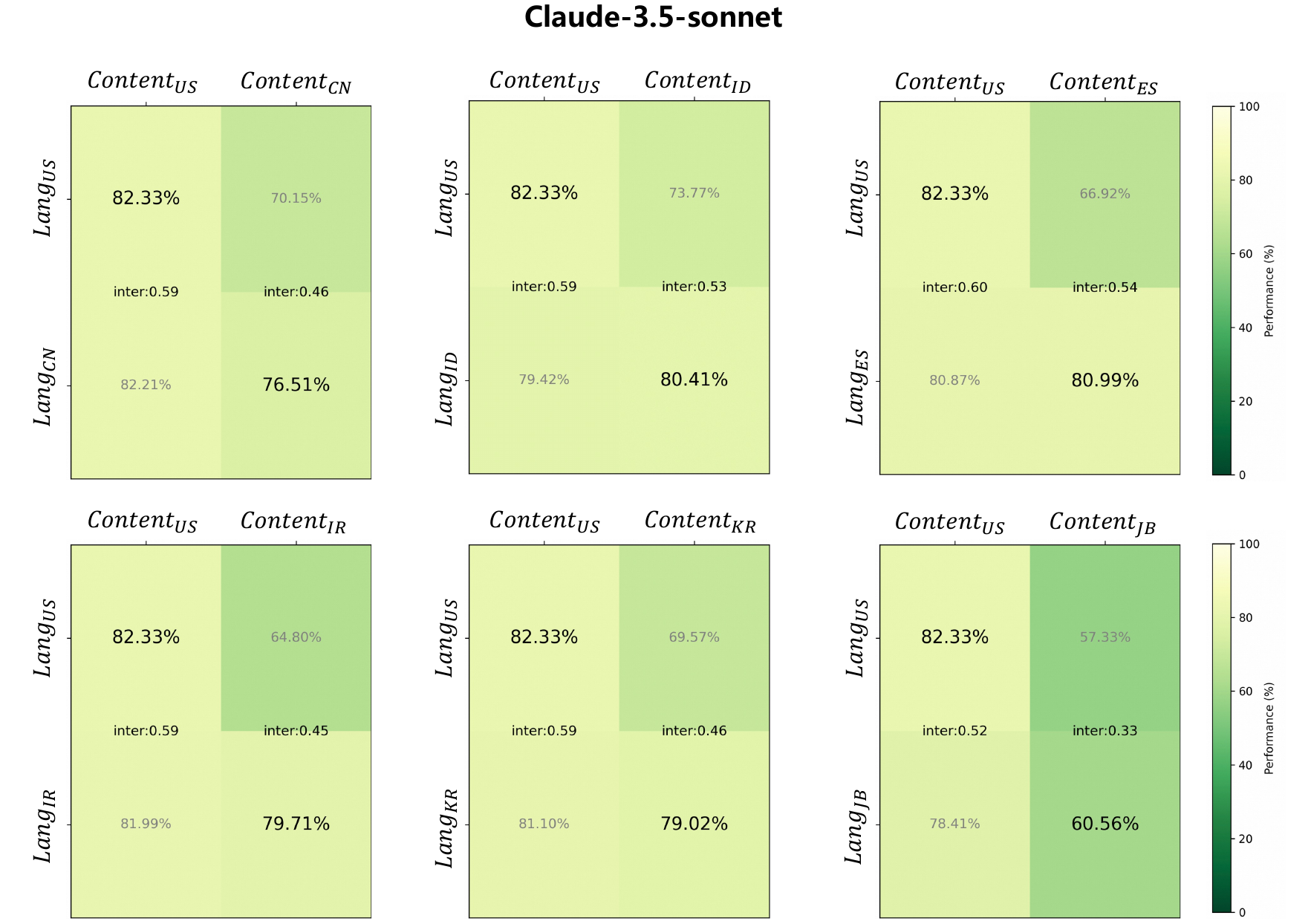}
    \caption{The average performance of Claude-3.5-sonnet on Instruction set~\ref{appen: instruction}. The $Content_{i}$ represents langue$i$-speaking culture context, $Lang_{i}$ represents the linguistic medium for language $i$.}
    \label{Claude-3-5}

     \vspace{10pt}
     \includegraphics[scale=0.42]{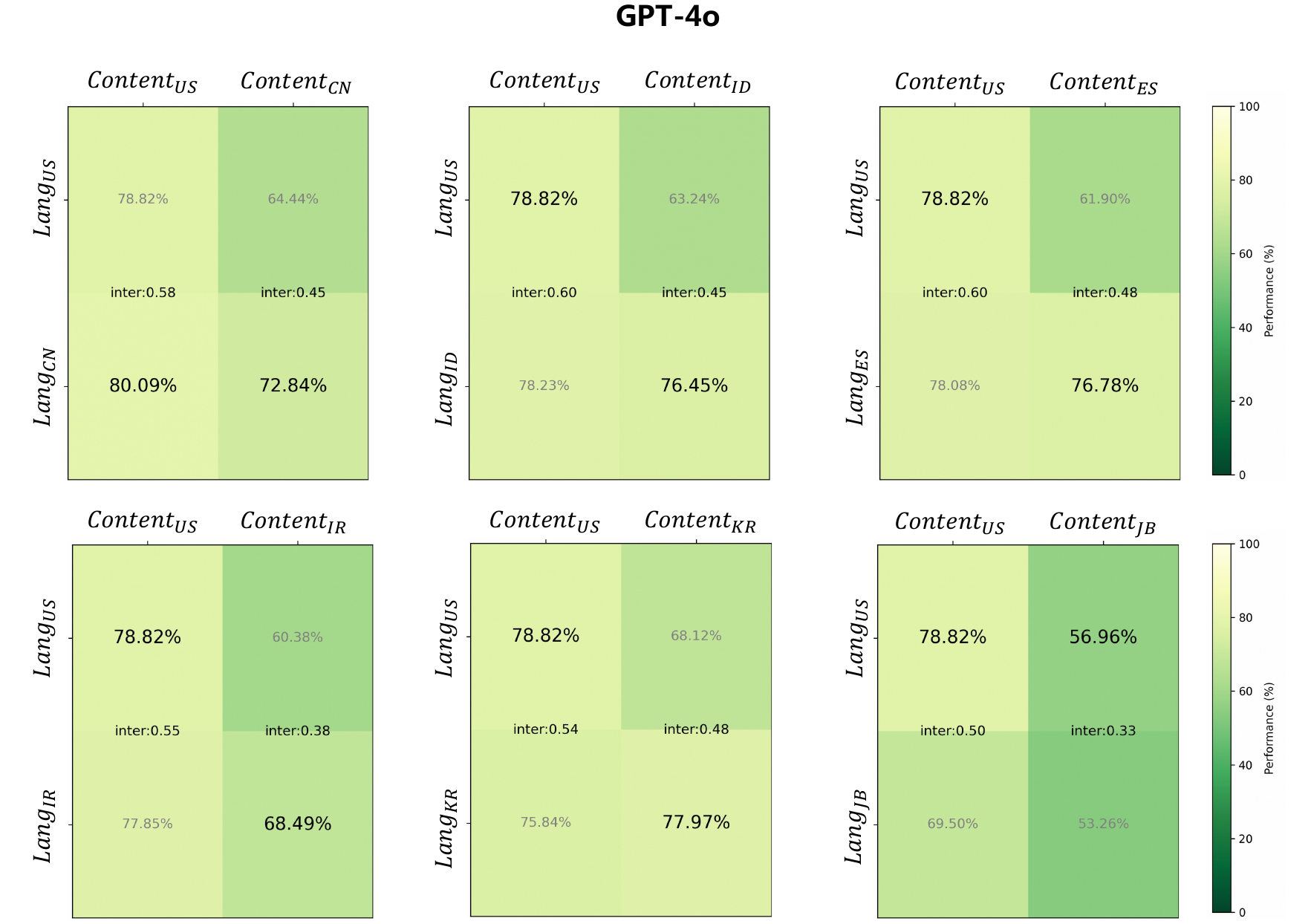}
    \caption{The average performance of GPT-4o on Instruction set~\ref{appen: instruction}. The $Content_{i}$ represents the langue$i$-speaking culture context, $Lang_{i}$ represents the linguistic medium for language $i$.}
    \label{gpt-4o}
   
\end{figure*}

\begin{figure*}[htb]
    \centering
    
    \includegraphics[scale=0.42]{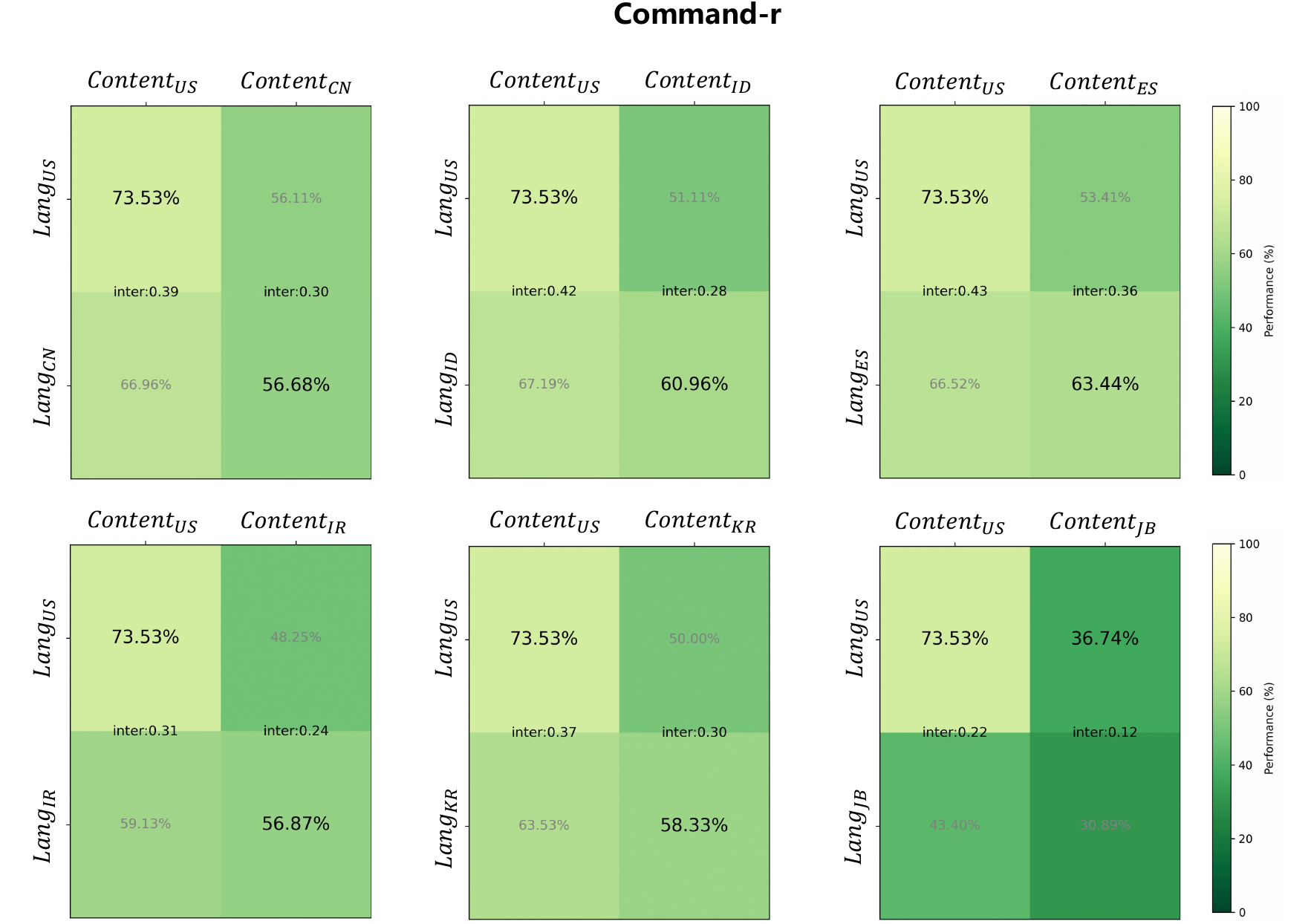}
    \caption{The average performance of Command-r on Instruction set~\ref{appen: instruction}. The $Content_{i}$ represents the langue$i$-speaking culture context, $Lang_{i}$ represents the linguistic medium  language.}
    \label{command-r}

     \vspace{10pt}
    
    \includegraphics[scale=0.42]{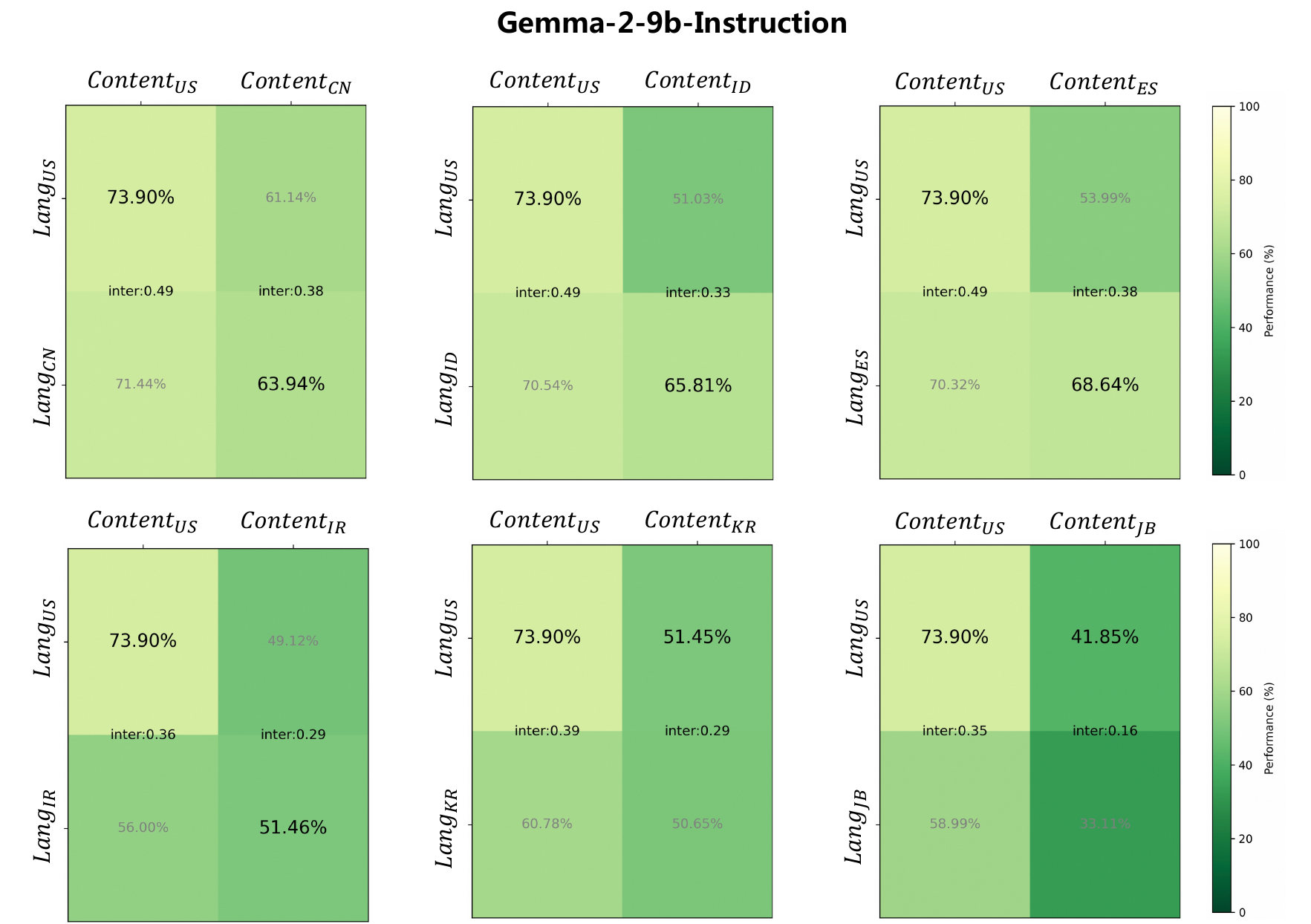}
    \caption{The average performance of Gemma-2-9b-Instruct on Instruction set~\ref{appen: instruction}. The $Content_{i}$ represents the langue$i$-speaking culture context, $Lang_{i}$ represents the linguistic medium for language $i$.}
    \label{gemma-2-9b-it}
\end{figure*}

\begin{figure*}[htb]
    \centering
    
    \includegraphics[scale=0.42]{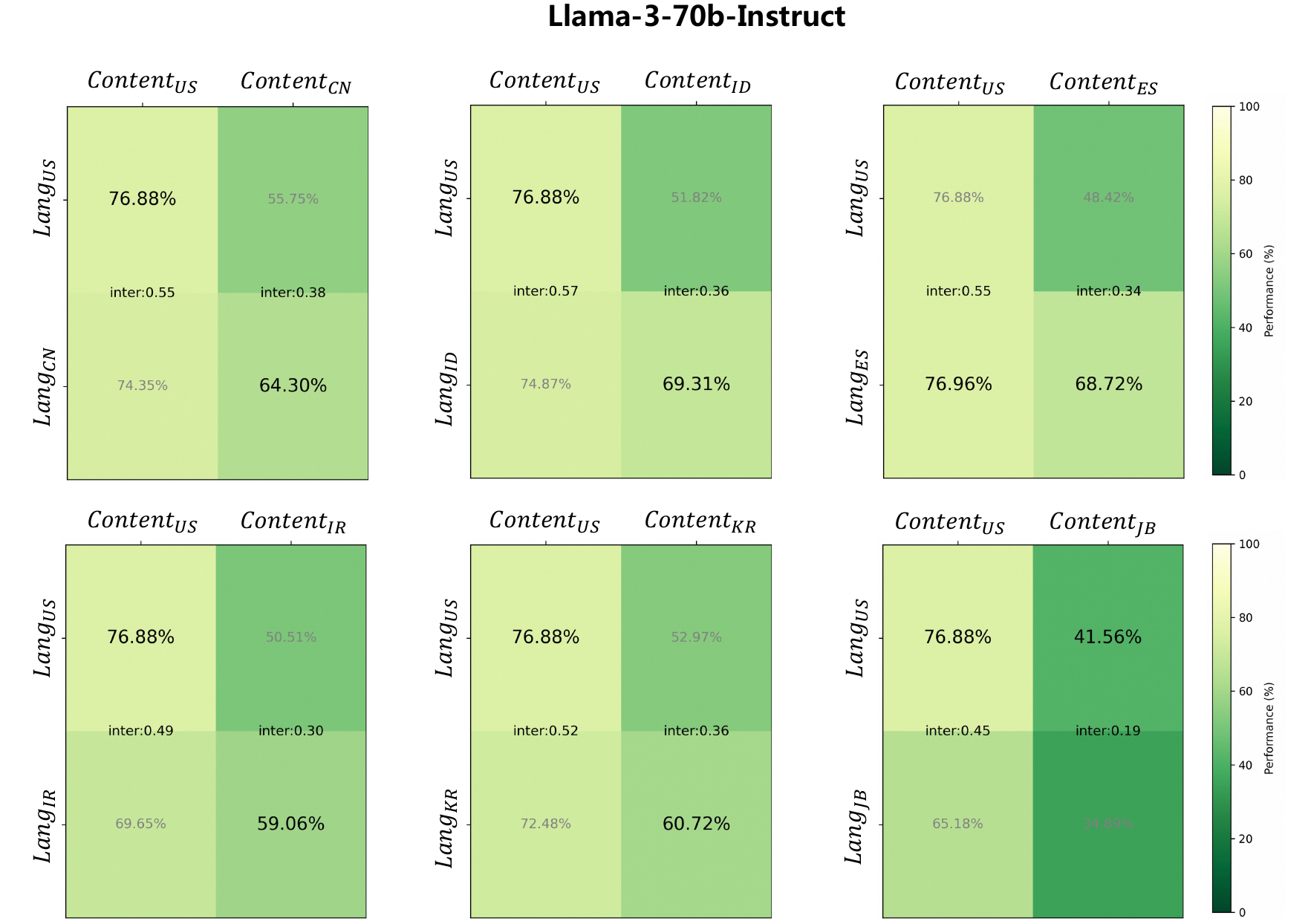}
    \caption{The average performance of Llama-3-70b-Instruct on Instruction set~\ref{appen: instruction}. The $Content_{i}$ represents the langue$i$-speaking culture context, $Lang_{i}$ represents the linguistic medium for language $i$.}
    \label{llama-3-70b}

     \vspace{10pt}
    
    \includegraphics[scale=0.42]{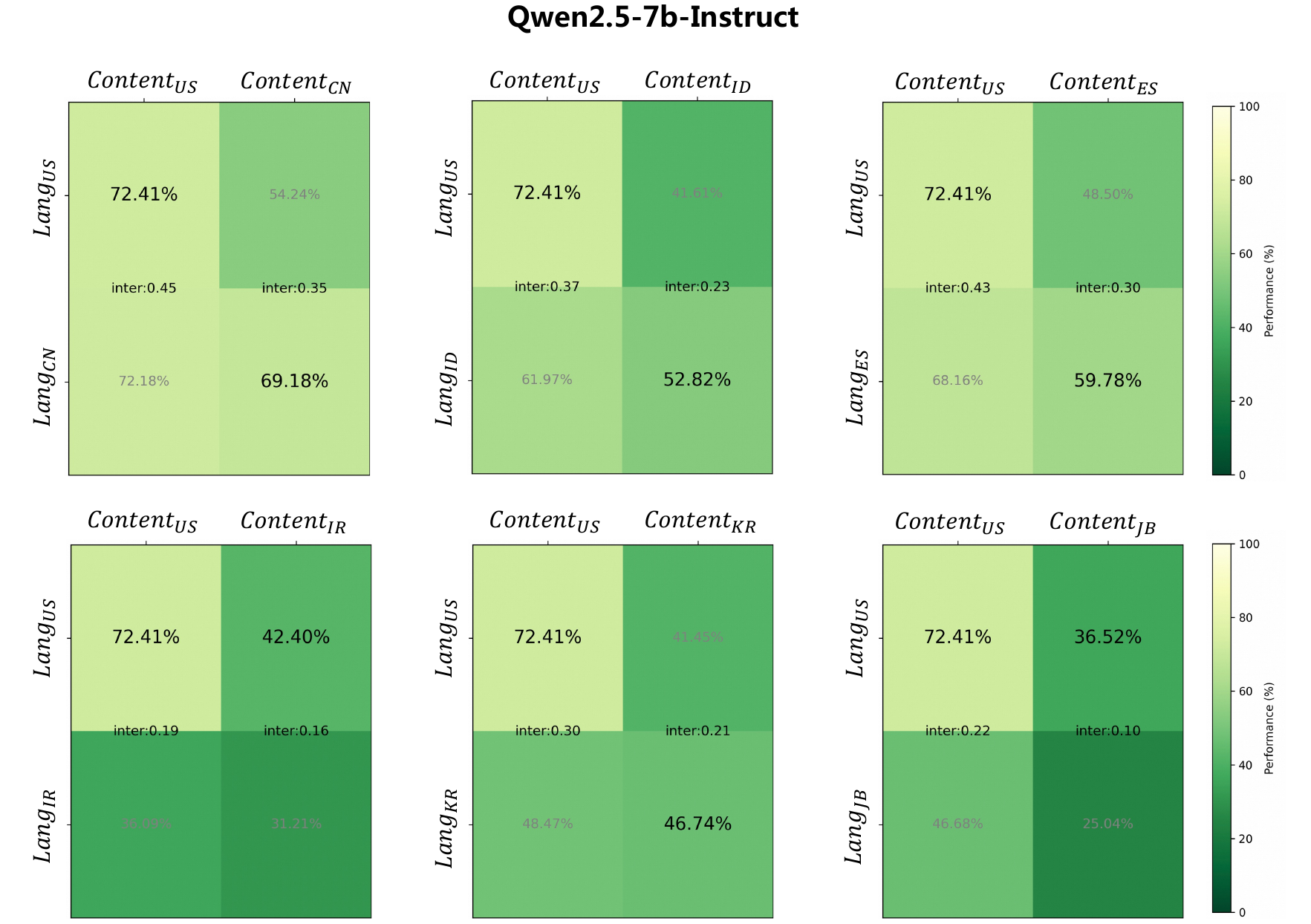}
    \caption{The average performance of Qwen-2.5-7b-Instruct on Instruction set~\ref{appen: instruction}. The $Content_{i}$ represents the langue$i$-speaking culture context, $Lang_{i}$ represents the linguistic medium for language $i$.}
    \label{Qw2.5}
\end{figure*}

\begin{figure*}[htb]
    \centering
    
    \includegraphics[scale=0.42]{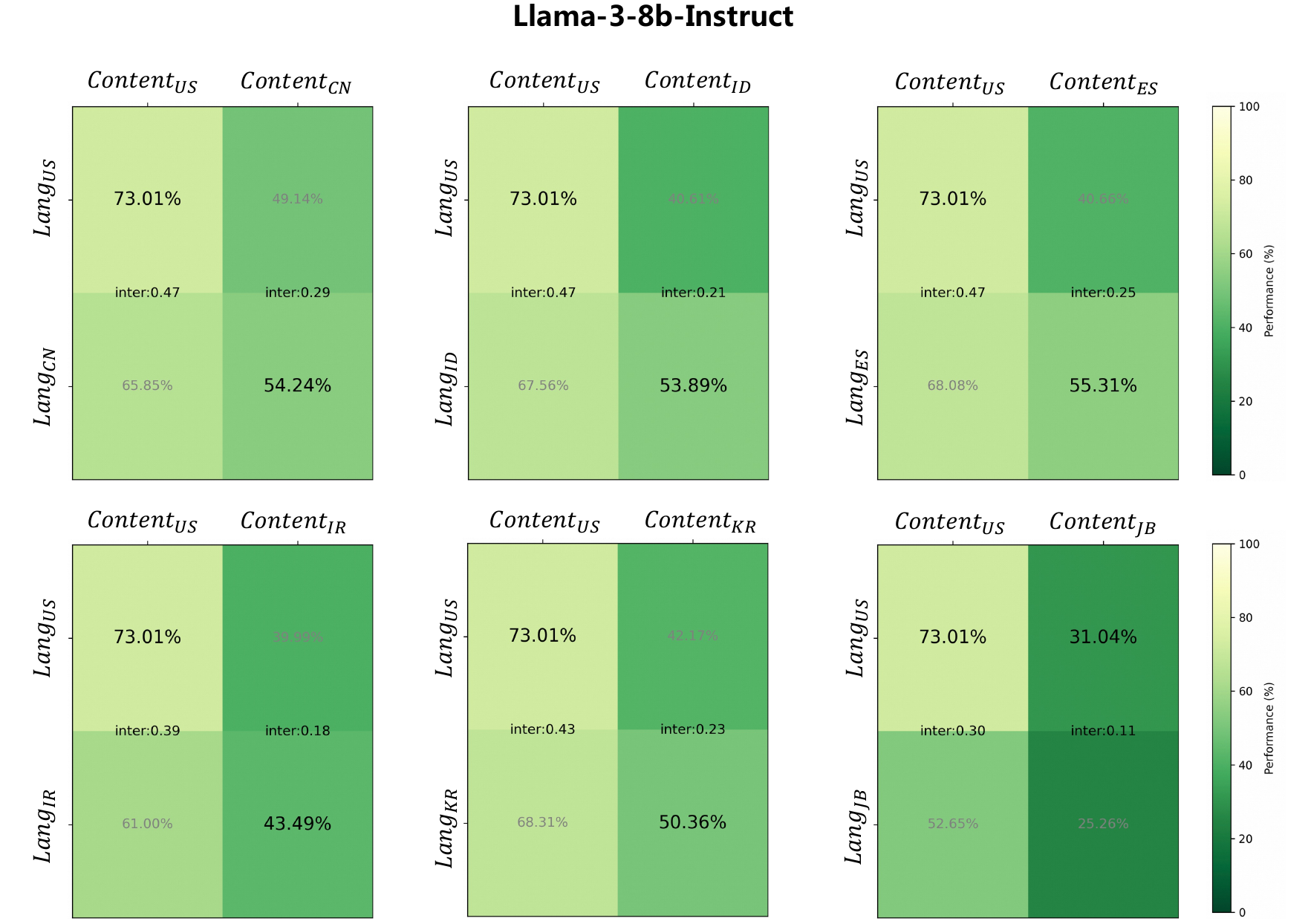}
    \caption{The average performance of Llama-3-8b-Instruct on Instruction set~\ref{appen: instruction}. The $Content_{i}$ represents the langue$i$-speaking culture context, $Lang_{i}$ represents the linguistic medium for language $i$.}
    \label{llama-3-8b}

     \vspace{10pt}
    
    \includegraphics[scale=0.42]{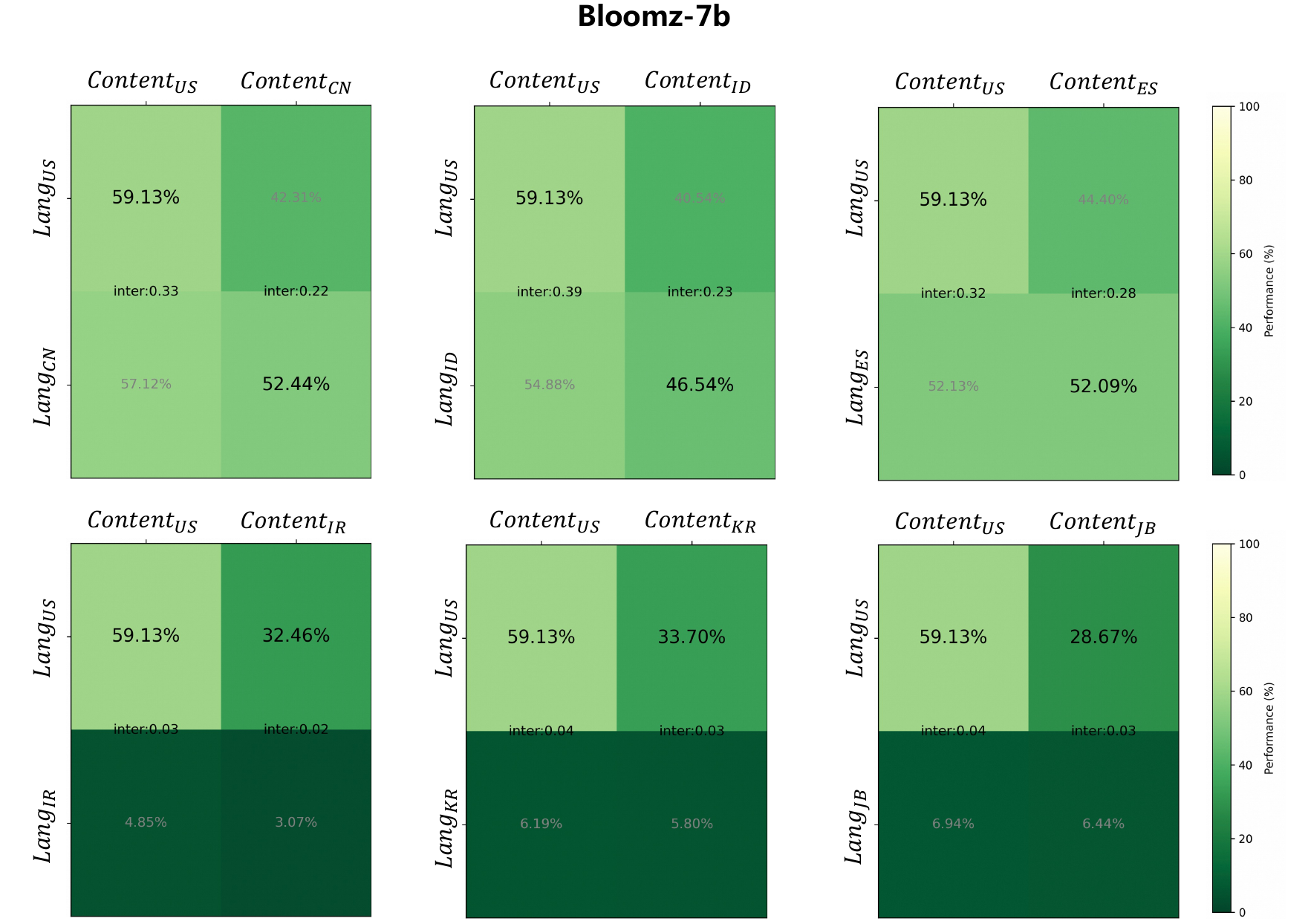}
    \caption{The average performance of Bloomz-7b on Instruction set~\ref{appen: instruction}. The $Content_{i}$ represents the culture context, $Lang_{i}$ represents the linguistic medium  language.}
    \label{bloomz}
\end{figure*}

\begin{figure*}[htb]
    \centering
    \includegraphics[scale=0.33]{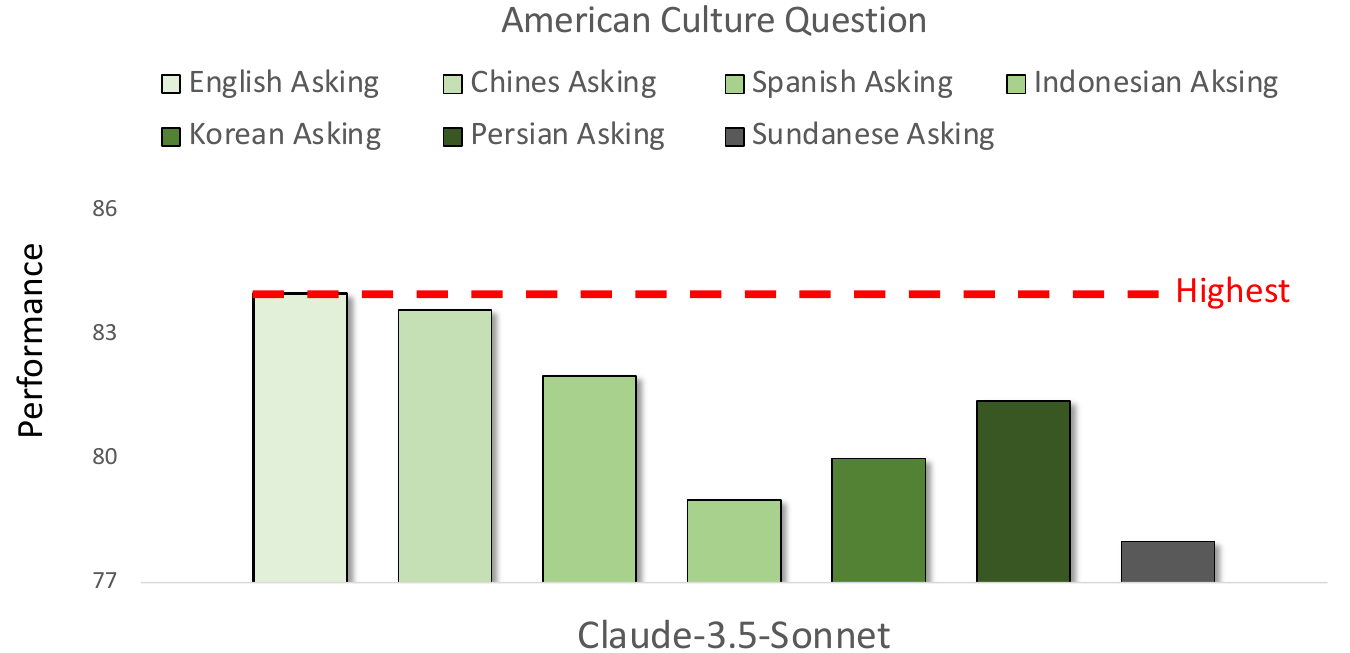}
    \caption{The performance of the selected models on the American culture question when asked in the other six languages versus English. Models perform the best when asking questions in English. }
    \label{fig: finding2_1}
\end{figure*}

The results presented in Section~\ref{sec: finding1} and Section~\ref{sec: finding2} are based on Instruction 1 (as shown in Appendix~\ref{appen: instruction}). All other results are displayed in an averaged format in Figures~\ref{Claude-3-5}, \ref{gpt-4o}, \ref{command-r}, \ref{gemma-2-9b-it}, \ref{llama-3-70b}, \ref{Qw2.5}, \ref{llama-3-8b}, \ref{bloomz}, and \ref{fig: finding2_1} for each model on every $Q_{i,j}$.

\begin{figure*}[htb]
    \centering
    \includegraphics[scale=0.35]{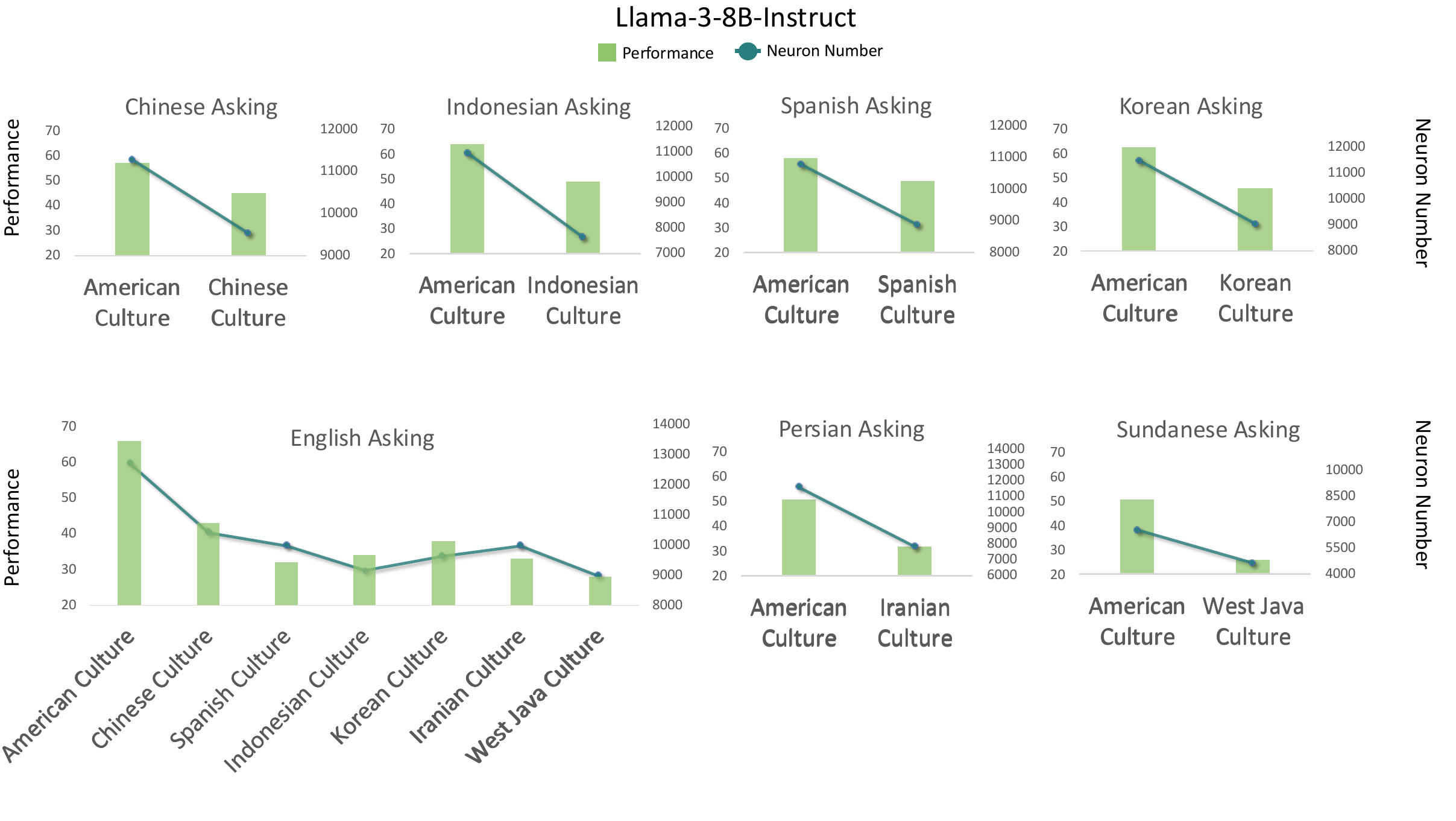}
    \caption{The performance and the number of Key Neurons for the Llama-3-8B-Instruction on cross-cultural contexts.}
    \label{fig: hypo3_all_llama3}
\end{figure*}

\begin{figure*}[htb]
    \centering
    \includegraphics[scale=0.35]{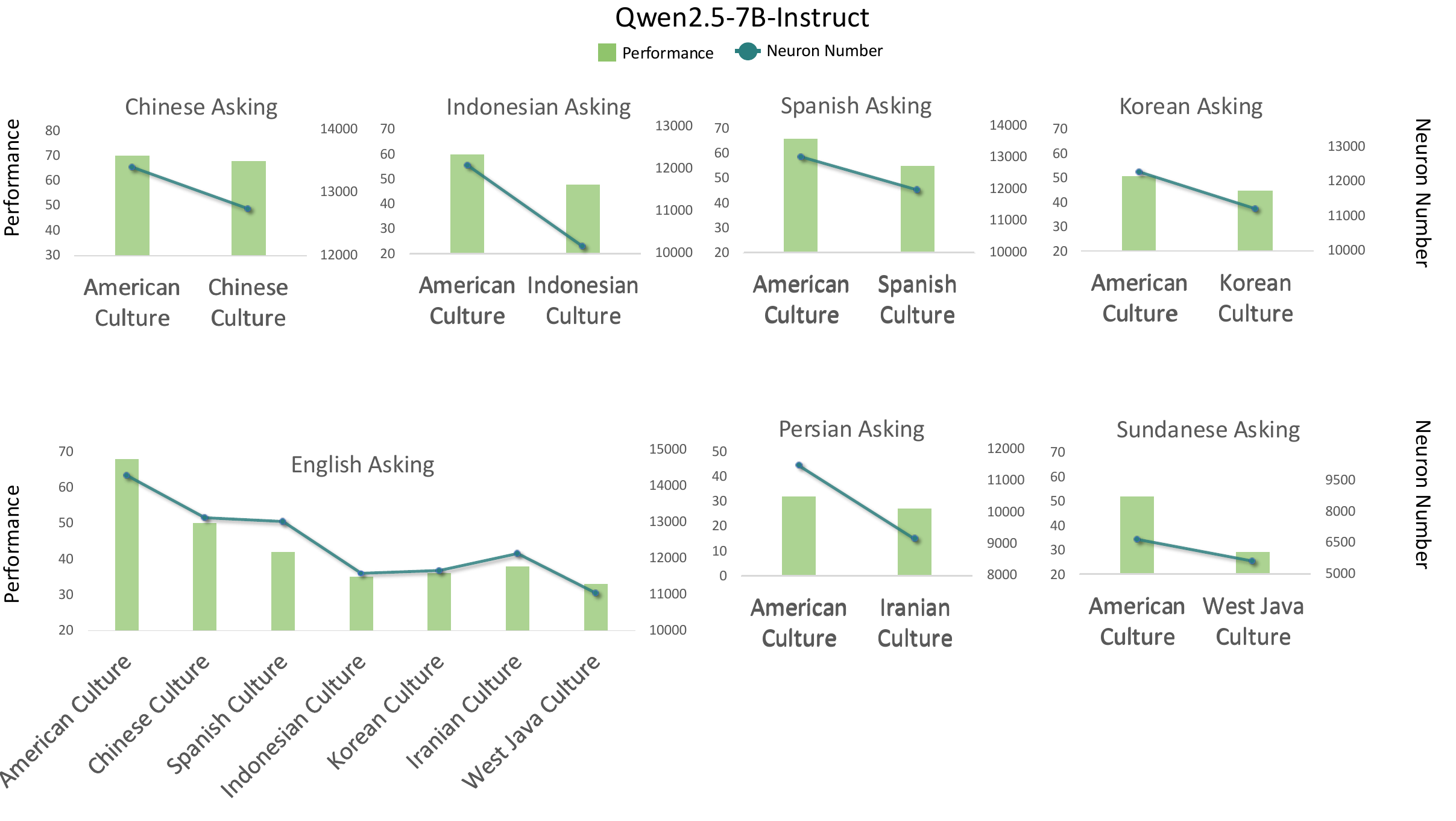}
    \caption{The performance and the number of Key Neurons for the Qwen-2.5-7B-Instruction on cross-cultural contexts.}
    \label{fig: hypo3_all_qwen}
\end{figure*}

For the results in Section~\ref{sec: hypo1}, Section~\ref{sec: hypo2}, and Section~\ref{sec: hypo3}, we conducted evaluations using Instruction 2 (as shown in Appendix~\ref{appen: instruction}) to manage computational costs. This is particularly relevant for certain questions where the model might generate lengthy responses, making the interpretation of results impractical without these adjustments.
For results in other languages, except English, which are not shown in Section~\ref{sec: hypo3}, please refer to Figures~\ref{fig: hypo3_all_llama3} and~\ref{fig: hypo3_all_qwen}.

\subsection{The primarily experiment result}
Table~\ref{tab: primarily} shows the performance of the models Llama-3-8B-Instruct, Gemma-2-9B-Instruct, and Qwen2.5-7B-Instruct on multilingual benchmarks: GSM8K and MMMLU. The experiment is conducted in a zero-shot setting, and the results suggest that the models perform better when the questions are asked in English compared to other languages.
\begin{table*}[htb]
\centering
\resizebox{0.8\textwidth}{!}{%
\begin{tabular}{lccc|ccc}
\toprule
\textbf{Model} & GSM8K$_{en}$ & GSM8K$_{cn}$ & GSM8K$_{es}$ & MMMLU$_{en}$ & MMMLU$_{id}$ & MMMLU$_{cn}$ \\
\midrule
Llama-3-8B-Instruct & \textbf{77.1} & 60.2 &66.7 &\textbf{64.4} &52.4 &54.5\\
Gemma-2-9B-Instruct & \textbf{81.2} & 77.9 &75.1 &\textbf{73.4} &64.4 &64.0\\
Qwen2.5-7B-Instruct & \textbf{84.3} & 80.3 &71.1 &\textbf{71.3} &56.8 &60.8\\
\bottomrule
\end{tabular}}
\caption{The performance for mode Llama-3-8B-Instruct, Gemma-2-9B-Instruct and Qwen2.5-7B-Instruct on   MMMLU~\cite{mmmlu}, MGSM~\cite{shi2022languagemodelsmultilingualchainofthought}. The experiment is conducted in a zero-shot setting. The languages we select are English(en), Chinese(cn), Spanish(es), and Indonesian(id). We find that models have better performance when the question is asked in English.}
\label{tab: primarily}

\end{table*}

\subsection{The completion for Dataset} \label{appen: completion for Dataset}
 
The original BLEnD includes, for each language-specific evaluation set $Q_{i,j}$ (except for English), an English translation evaluation dataset $Q_{i,en}$. For the rest language $(i, j)$ pair (when $i = en$), we deploy GPT-4o to conduct the translation. To ensure the translated question $q_{en,i}$ aligns  $q_{i,i}$ with the dual-format structure, we prompt GPT-4 with a one-shot example using the question pair  $q_{i,en}$, $q_{i,i}$ to obtain the translated version $q_{en,en}$  for language  $i$ , which we then use as  $q_{en,i}$.

To further evaluate the quality of this complement, we conduct a human evaluation involving four senior computational linguistics researchers who have a research focus in multilingualism and are trained in advanced. From the constructed $Q_{en,i}$, we randomly sampled 100 cases, along with their dual cases from $Q_{en,en}$, $Q_{i,en}$, and $Q_{en,en}$. We presented these 100 paired cases ($q_{i,en}$, $q_{i,i}$, $q_{en,en}$, and GPT-o translated $q_{en,i}$) to the evaluators, asking them to score the translated content and format consistency: 1 point: The translation content is problematic or inaccurately expressed. 2 points: The translation content is accurate, but the format deviates significantly from the corresponding $q_{i,i}$. 3 points: The translation content is accurate, and the format aligns perfectly with the corresponding $q_{i,i}$. The results indicate that the average full mark rate (3 points) for translated content and format consistency is 97.8\%, with scores above 2 points reaching 100\%. The overall agreement rate is 95\%. This prove the quality of the newly introduced dataset.

\subsection{The threshold function for Interpretation} ~\label{appen: thresholdselection}

In our experiment, we deploy top-k (k = 5) threshold for each layer. Specifically, we compute the activation value for each corresponding response token $r_i$ for each question $q$. We then aggregate the activation scores of each neuron for each response token across each layer $l$ ( $l \in \{ 1, 2, \ldots, L \} $ ), represented as: $ V_{l} = \left[ v_{(j, l)}^{r_i} \mid r_i \in r, j \in \left\{1, 2, \ldots, dm\right\} \right] $. To determine the key neurons for question $q$, we select the top-k neurons for each layer, forming the key neuron set $N_q$ as: 

\begin{align}
N_q = &\big\{\, (j, l) \;\big|\;  v_{(j, l)}^{r_i} \geq V_{l}^{\text{top-}k}, r_i \in r,\  \nonumber \\
& j \in \{1, 2, \ldots, dm\},\ l \in \{1, 2, \ldots, L\} \,\big\}.
\end{align}

When we conduct experiments, we also explore other threshold function. Including:  1) Layer-specific top-k (final adoption in the paper) 2) Global top-k, 3) Global top-k score,
4) Global top-k score.
 We determine the threshold by conducting the experiment shown in Section~\ref{sec: abla} and select the optimal ones.

\section{Instruction} \label{appen: instruction}
We mainly use the instructions from the original benchmark BlEnD~\cite{NEURIPS2024_8eb88844}. However, some models’ responses are longer due to the nature of the instruction, so to better match each question with candidate answers and help us conduct the interpretation experiment, we manually add additional instructions (instruction 2 for each language). 

\begin{figure*}[htb]
    \centering
    \includegraphics[scale=0.72]{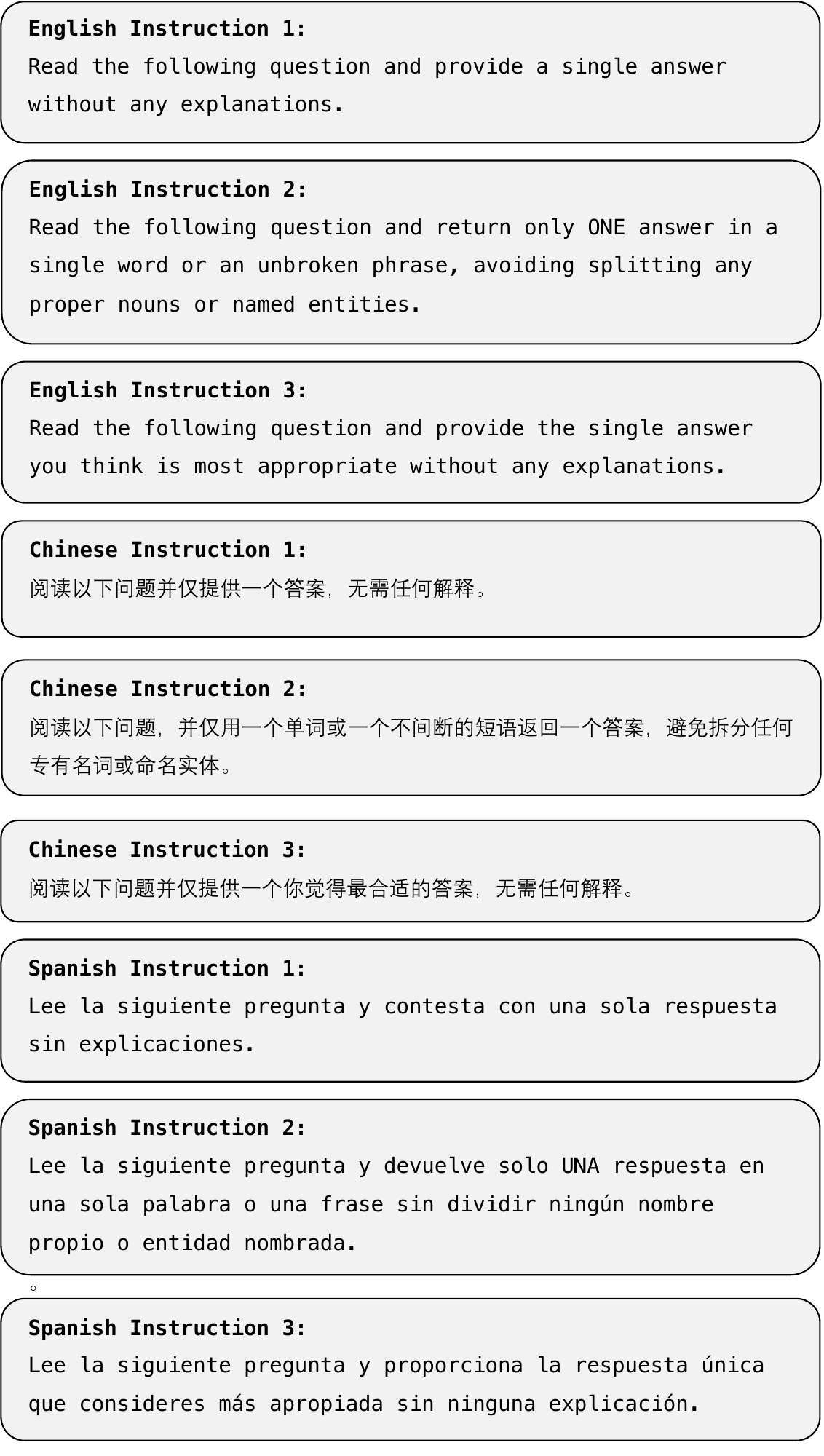}
    \captionsetup{labelformat=empty} 
\end{figure*}
\begin{figure*}[htb]
    \centering
    \includegraphics[scale=0.72]{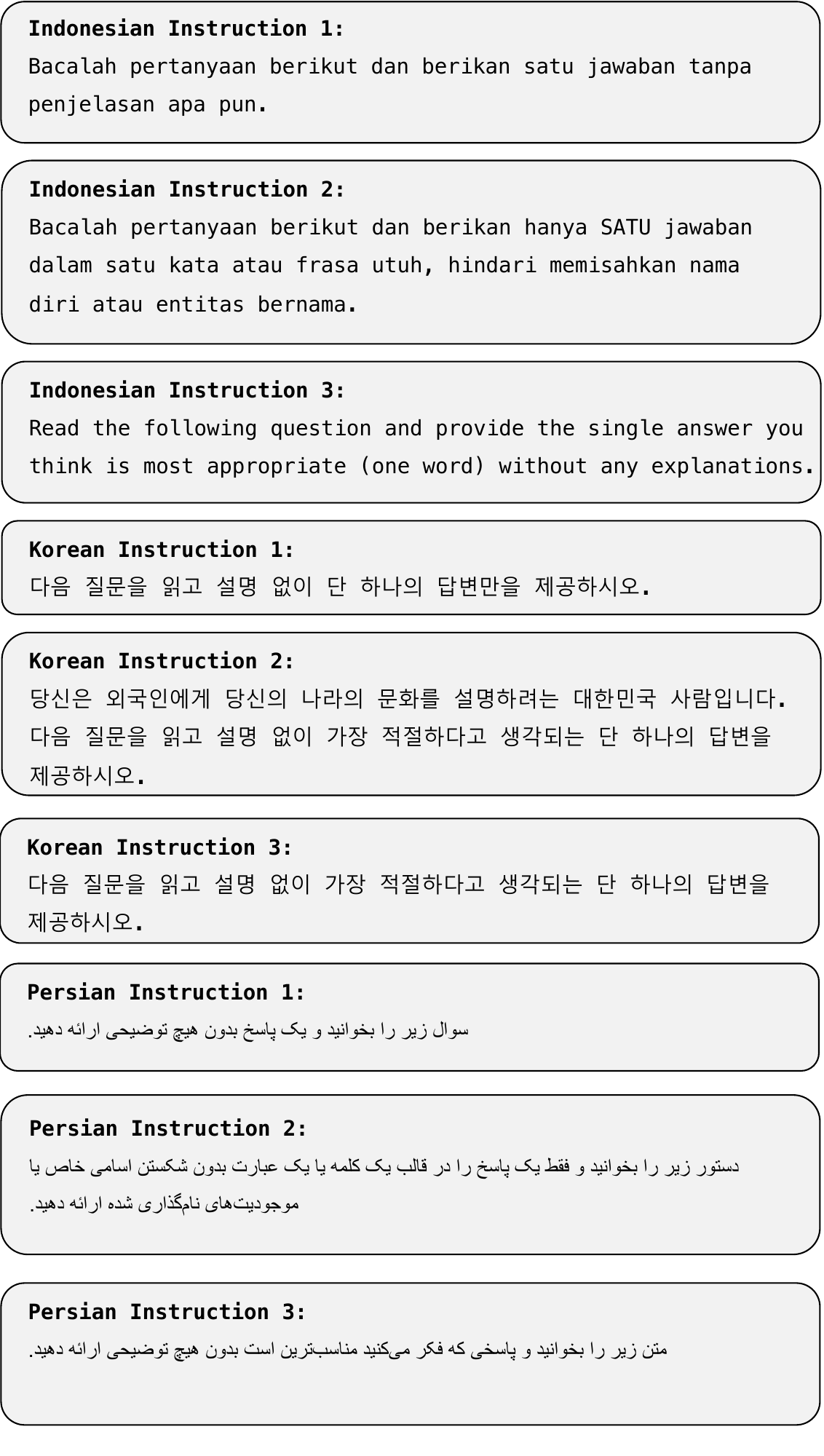}
    \captionsetup{labelformat=empty} 
\end{figure*}
\begin{figure*}[htb]
    \centering
    \includegraphics[scale=0.72]{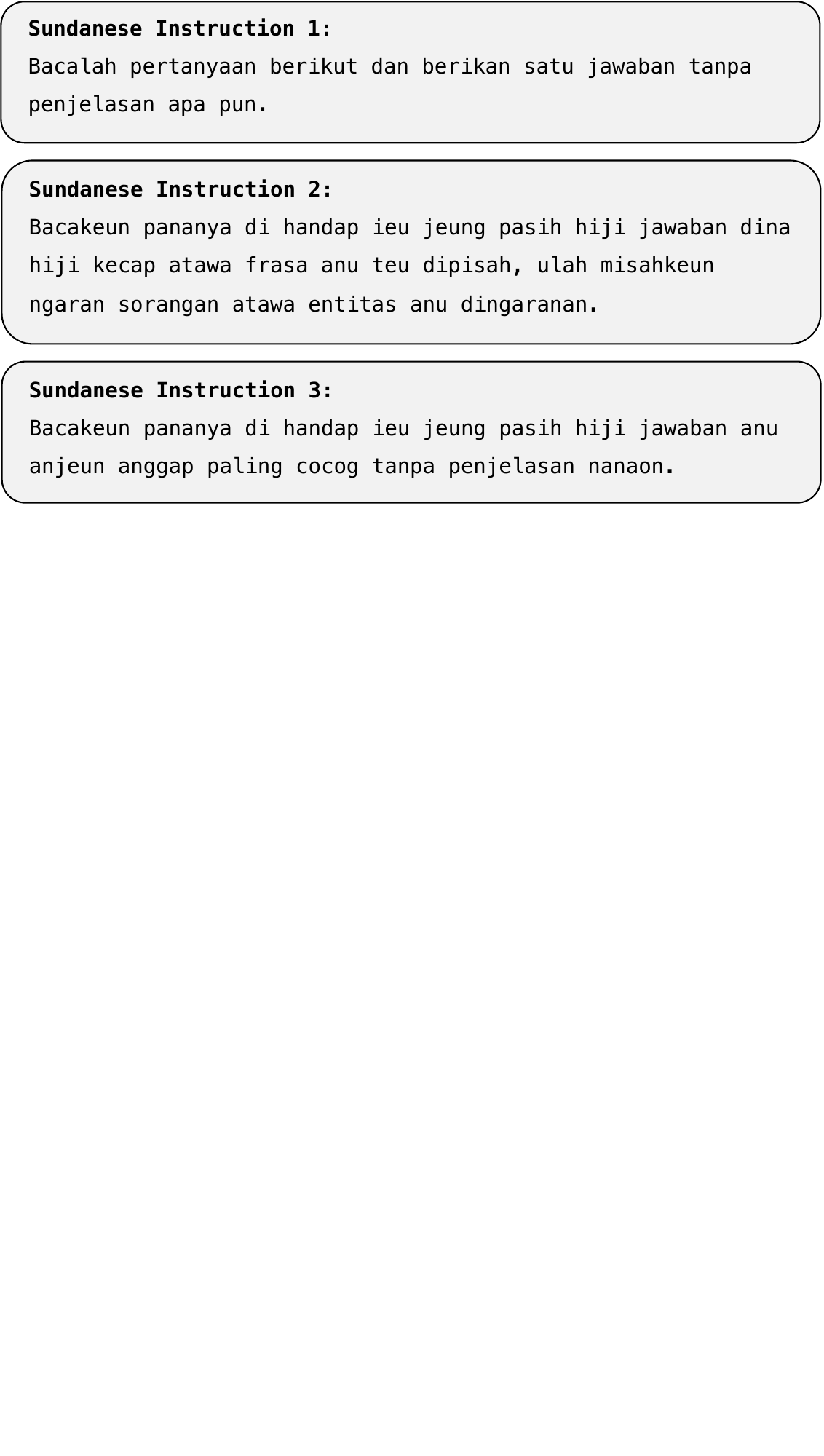}
    \captionsetup{labelformat=empty} 
\end{figure*}

\end{document}